%% file: iclr2027_conference.tex
\documentclass{article}
\usepackage{iclr2027_conference,times}

\input{math_commands.tex}

\usepackage{amsmath}
\usepackage{amssymb}
\usepackage{xcolor}
\usepackage{hyperref}
\usepackage{url}
\usepackage{graphicx}
\usepackage{float}
\usepackage{wrapfig}
\usepackage{colortbl}   
\usepackage{array}
\usepackage{booktabs}
\usepackage{enumitem}
\usepackage{fvextra}
\usepackage{tcolorbox}
\tcbuselibrary{listings,breakable,skins}

\title{Internalizing Agent Experience into Diffusion Model Weights via On-Policy Context Distillation}

\author{%
\parbox{0.95\textwidth}{\centering
{\small\mbox{Wenxuan Wang$^{1,2,*}$ \quad
Zekai Liu$^{3,*}$ \quad
Weinan Zhang$^1$ \quad
Yu Cheng$^{4,\dagger}$ \quad
Yang Yang$^{5,\dagger}$}}\\[6pt]
{\normalfont\small
\mbox{$^1$Harbin Institute of Technology \quad
$^2$Shanghai AI Laboratory \quad
$^3$Shandong University}\\
\mbox{$^4$Nanyang Technological University \quad
$^5$Shanghai Jiao Tong University}\\[3pt]
$^*$Equal contribution. \quad
$^\dagger$Corresponding author.
}}}

\iclrfinalcopy
\begin{document}

\maketitle
\lhead{Preprint}
\begin{abstract}
Wrapping an image generation model in an agentic harness can effectively boost
Text-to-Image task performance: the harness can leverage memory, skills, workflow
orchestration, result verification, and iterative refinement to continually construct
and revise prompts, thereby eliciting better images. These gains, however, remain
external to the diffusion model and are realized only while the full harness runs.
We propose Diffusion On-Policy Context Distillation (D-OPCD), which treats the
agent-improved prompt as privileged context and distills the knowledge encoded
in the agent harness into the weights of the diffusion model, so that the model
retains part of the harness's benefit when conditioned on the original query alone.
Using a Text-to-Image agent equipped with our proposed Auto Skill Evolver
(ASE), we show that D-OPCD can internalize harness capabilities into the
generator's weights, raising the average direct-generation score from 60.52 to
65.09 across four benchmarks.
With this knowledge absorbed into the weights, the harness can shed its saturated skills and resume evolving: a second ASE round on the updated generator improves on a skill-free harness by additional 1.83 points, pointing toward text-to-image systems in which harness and model keep improving each other through continual co-evolution. \href{https://github.com/Yummytanmo/D-OPCD-CoEvolution}{Code} is publicly available.

\end{abstract}

\section{Introduction}

Agentic harnesses have become an effective way to improve text-to-image
(T2I) generation. A harness wraps a fixed generator with reasoning,
memory, skills, and verification: within a task, it reasons about and
iteratively refines its generations~\citep{wang2025imagent,he2026gems};
across tasks, it accumulates experience so that useful strategies carry
over to later requests~\citep{chen2026memogen}. As the harness evolves,
it learns to use the generator better. The generator itself, however,
does not change: the acquired gains live in external guidance that must
be retrieved and applied again on every subsequent request.

Keeping these gains outside the generator has two costs. First, as memories and skills accumulate, relevant guidance becomes harder to retrieve. Second, reapplying this guidance to every request incurs repeated reasoning, verification, and image generation. Internalizing established gains could reduce both costs: a one-time offline training cost is amortized over future requests, and obsolete strategies can be discarded so the harness can target the updated generator’s remaining limitations. This raises our central question: can gains discovered through harness evolution be internalized into the generator, enabling it to retain these improvements without external guidance at inference time?

Answering this question requires an interface between harness experience and generator learning. Although agents represent experience through different memories, skills, and reasoning traces, all ultimately condition the T2I generator on a text prompt. We therefore retain, for each completed task, the original query and the harness-generated prompt that yielded the selected image. This captures prompt-mediated gains, but not gains from verification or stochastic sample selection.
Prior work uses visual experience to improve the agent policy that constructs generator inputs~\citep{chen2026genevolve}, leaving the harness responsible for realizing these gains at inference time. In contrast, we use the resulting query--prompt pairs to train the generator itself, making the gains available from the original query alone.

To perform this transfer, we introduce \emph{Diffusion On-Policy Context Distillation (D-OPCD)}, a self-distillation algorithm for diffusion models. A teacher conditioned on both the original query and the harness-produced prompt supervises a student that receives only the original query, at states sampled from the student’s own denoising trajectories. We formulate harness-to-generator transfer as on-policy context distillation~\citep{ye2026opcd} for diffusion models, with the evolving harness providing privileged textual context. Unlike prior on-policy diffusion self-distillation~\citep{jiang2026dopsd}, which obtains privileged information from target or reference images, D-OPCD distills the gains elicited by the harness. It thereby internalizes prompt-mediated harness gains into the generator’s weights, making them available without the harness at inference time.

We study this process with Auto Skill Evolver (ASE), a test-time learning mechanism that turns completed tasks into reusable prompt guidance; the resulting query--prompt pairs train D-OPCD. Distilling these pairs improves the generator's direct-generation performance across all four benchmarks. With the skill-free harness reattached, the updated generator outperforms its base-generator counterpart and even slightly exceeds the original skill-equipped agent on average, indicating that its learned skills have been internalized. Crucially, this establishes an iterative improvement cycle: after the generator internalizes the learned skills, resetting those now-redundant skills allows ASE to evolve again, yielding further improvements.

In summary, our main contributions are as follows:
\begin{itemize}[label=\textbullet,align=left,leftmargin=1.2em,labelindent=0pt,labelwidth=0.6em,labelsep=0.6em,topsep=4pt,itemsep=4pt,parsep=0pt]
\item We introduce D-OPCD, an on-policy context distillation algorithm that uses harness-produced prompts as privileged teacher context to internalize prompt-mediated gains into diffusion model weights, making part of the harness's benefit available from the original query alone.
\item We develop Auto Skill Evolver (ASE), a test-time learning mechanism that turns task experience into reusable skills and provides query--prompt pairs for D-OPCD. Experiments across four benchmarks demonstrate improved direct generation after internalization.
\item We formulate harness--model co-evolution for T2I agents by coupling harness adaptation with generator learning. Internalizing accumulated harness gains into the generator allows the harness to shed its saturated skills and resume evolving around the updated model.
\end{itemize}

\section{Related Work}
\label{sec:related_work}

\subsection{Agentic Text-to-Image Generation}

The quality of a generated image depends in part on how a user's request is translated into instructions for the generator. Prompt adaptation improves this interface by rewriting requests into prompts better suited to a fixed T2I model~\citep{hao2023optimizingpromptstexttoimagegeneration}. For more complex requests, image-generation agents combine planning, generation, visual inspection, and prompt revision, using feedback from earlier attempts to address missing objects or unsatisfied constraints~\citep{wang2025imagent,jiang2026genagent,kovalev2025craftcontinuousreasoningagentic,zhang2026qwenimageagent}. This process can also extend beyond a single task: GEMS equips the agent with memory and skills, MemoGen reuses experience from earlier requests, and GenEvolve learns from generation trajectories to improve the agent's decisions~\citep{he2026gems,chen2026memogen,chen2026genevolve}. These approaches show how better prompt construction and accumulated experience can improve the use of an image generator over time.

\subsection{Training Diffusion Models}

Updating a diffusion generator's weights provides another route to improving its outputs. Supervised fine-tuning can teach new visual concepts from a small set of examples, as in DreamBooth, or support continual customization across concepts~\citep{ruiz2023dreamboothfinetuningtexttoimage,smith2023continualdiffusion}. For qualities that are harder to specify with target images, training can instead use evaluative signals: ImageReward supplies preference-based feedback for ReFL, DDPO optimizes rewards over denoising trajectories, DRaFT backpropagates differentiable rewards through sampling, and Diffusion-DPO learns from preferred image pairs~\citep{Xu_2023,black2024trainingdiffusionmodelsreinforcement,clark2024draft,Wallace_2024}. Recent methods extend online reward optimization to flow-matching models and use teacher predictions along a student's own sampling trajectory to improve diffusion or flow generators~\citep{liu2025flowgrpo,fang2026flowopd,jiang2026dopsd}. This literature offers several ways to turn examples or feedback into persistent changes in generation behavior.

\subsection{Harness--Model Co-Evolution}
\label{sec:rw_coevolution}

Experience-driven agent improvement begins by recording what succeeded or failed and turning those observations into guidance for later tasks. ExpeL extracts reusable lessons from trajectories, while MemSkill learns and revises skills for managing an agent's memory~\citep{Zhao_2024,zhang2026memskill}. Recent work connects this external adaptation to policy learning by evolving skills alongside reinforcement learning~\citep{xia2026skillrl,zhang2026coevolvingskillgenerationpolicy}, updating an agent's harness and model across interactions~\citep{xia2026metaclaw,karten2026continualharness}, or deriving hindsight skills that supervise the current policy~\citep{wu2026seedselfevolvingonpolicydistillation}. Across these settings, the current model shapes the experience available for the next update, while accumulated experience changes the model and the guidance surrounding it. We study this interaction in T2I generation, where the evolving harness constructs prompts for a separate image generator.

\begin{figure*}[t]
    \centering
    \includegraphics[width=\textwidth]{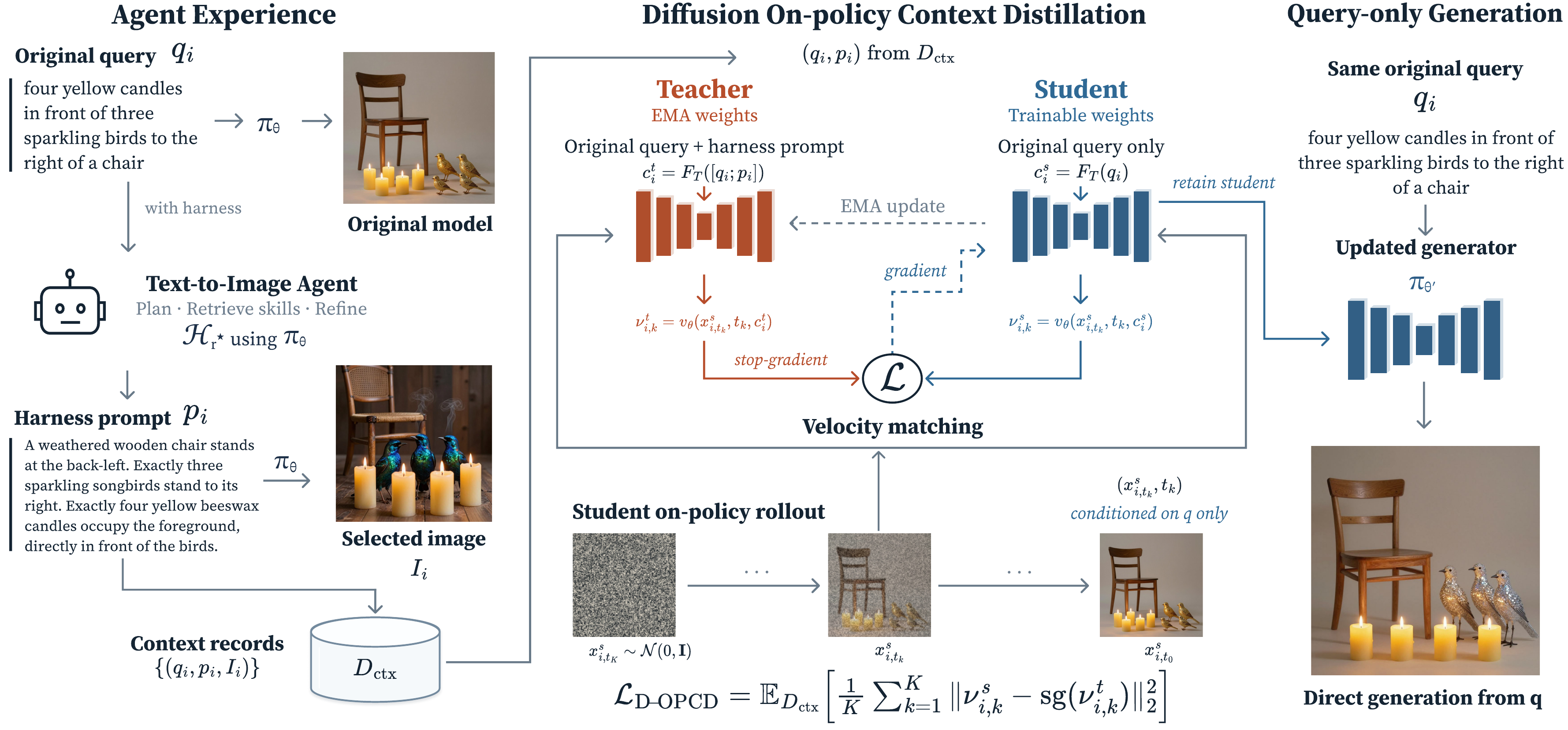}
    \caption{\textbf{Method overview.} As tasks accumulate, a T2I harness
    learns reusable skills that improve its prompts while the generator remains
    fixed. D-OPCD conditions an EMA teacher on the original query $q_i$ and the
    harness-selected prompt $p_i$, and trains a student conditioned only on
    $q_i$ to match the teacher's predictions along the student's own denoising
    trajectory. The updated generator can generate from $q_i$ alone, while the
    harness resets its skills to continue adapting around it.}
    \vspace{-12pt}
    \label{fig:method_overview}
\end{figure*}

\section{Method}
\subsection{Harness Gains at the Generator Interface}
In a training-free T2I agent in which the harness may invoke the generator multiple times, the final performance gains arise mainly from two main channels: improving the conditioning supplied to a frozen diffusion model and using external tools to compensate for its limitations. Harness components—including memory, skills, planning, verification, and reflection—may interact in harness-specific ways to realize these gains. We focus on the prompt-mediated gain exposed at the generator interface: for a task with its original query $q_i$, each invocation of the diffusion model produces a prompt–image pair,
\begin{equation}
\mathcal{G}_i
=
\{(p_{i,j},I_{i,j})\}_{j=1}^{J_i},
\qquad
I_{i,j}\sim\pi_\theta(\cdot\mid p_{i,j})
\end{equation}
where each $p_{i,j}$ records a generation
condition constructed by the harness and \(J_i\) is the number of diffusion model generation made by the harness for task $i$.
Let $j_i^\star$ denote the call selected by the harness for submission, and define
\begin{equation}
(p_i,I_i)
:=
(p_{i,j_i^\star},I_{i,j_i^\star})
\in\mathcal G_i.
\label{eq:selected-pair}
\end{equation}

Under a fixed generator and sampling protocol, the quality of $I_i$ provides an observable proxy for the effectiveness of $p_i$. The selected pair therefore provides a design-agnostic interface for internalization: $p_i$ encodes the condition constructed by the harness, while $I_i$ records its realized outcome. 

\subsection{Diffusion On Policy Context Distillation}
\label{sec:dopcd}
\subsubsection{Harness Internalization as Context Transfer}
\label{sec:harness_context_transfer}
The selected prompt $p_i$ is constructed and retained by the harness after
reasoning, refinement, and verification. It therefore serves as a task-specific
compilation of the harness's prompt-mediated experience, while the associated
image $I_i$ provides evidence that this condition is effective. Our goal is
not merely to reproduce $I_i$ from $p_i$, but to transfer the generation
advantage \textit{encoded} in $p_i$ to the model when it receives only the original
query $q_i$.

Direct SFT cannot realize this transfer. Training on $(p_i,I_i)$ leaves the harness knowledge in the input and does not teach query-only generation, whereas training on $(q_i,I_i)$ discards this privileged signal and supervises image-derived rather than student-visited states, creating an off-policy mismatch. We therefore introduce \emph{Diffusion On-Policy Context
Distillation} (D-OPCD), which the teacher
has access to privileged context that is hidden from the student ~\citep{ye2026opcd}.

Specifically, D-OPCD uses a teacher diffusion model conditioned on both $q_i$
and $p_i$ to supervise a student model conditioned only on $q_i$, with both evaluated along the student’s own
denoising trajectory, following
the on-policy diffusion construction~\citep{jiang2026dopsd}. This provides a training signal without additional evaluator queries and directly
targets the transfer from harness-assisted generation to generation under the original query.

\subsubsection{On-Policy Distillation Objective}
\paragraph{Harness version and context dataset.}
Let $\mathcal{H}_r$ denote a snapshot of the harness, where $r$ indexes its version. The harness may evolve through self-evolution
(Section~\ref{sec:self-evolve}) or human tuning, producing successive versions.
A version $\mathcal{H}_r$ is admitted for internalization only if it improves over query-only generation on the held-out task set $\mathcal{Q}_{\mathrm{hold}}$
(see the data protocol in Section~\ref{sec:coevolution}). Let $\pi_{\theta}$ denote the generator on
which $\mathcal{H}_r$ operates. This generator stays frozen while
$\mathcal{H}_r$ produces data and is modified only by the D-OPCD update described below.

Running $\mathcal{H}_r$ with $\pi_{\theta}$ on $\mathcal{Q}_{\mathrm{train}}$ produces
completed tasks. Each task has a submitted pair $(p_i,I_i)$, defined in
\Eqref{eq:selected-pair}. Recall that the harness selects $p_i$ itself: $p_i$ is the final generation condition whose realized image passed the harness's internal verification.
We retain $N_r$ completed-task records to form the context dataset for $\mathcal{H}_r$:
\begin{equation}
\mathcal{D}_{\mathrm{ctx}}^{(r)}
=
\{(q_i,p_i,I_i)\}_{i=1}^{N_r}.
\label{eq:dctx}
\end{equation}
The image $I_i$ is used only for record selection, where it filters out tasks whose submitted outputs fail the selection criteria. D-OPCD only uses $p_i$ as the sole privileged context.

\paragraph{Asymmetric conditioning.}
The text-conditioning encoder $F_{\mathrm{T}}$ maps text into the generator's conditioning space. We define two branches:
\begin{equation}
c_i^{s}
=
F_{\mathrm{T}}(q_i),
\qquad
c_i^{t}
=
F_{\mathrm{T}}([q_i;p_i]),
\end{equation}
where $[q_i;p_i]$ denotes a structured text condition consisting of the
original query followed by the harness-selected prompt.
The student receives only the original user condition. The teacher receives
the same condition together with the context constructed by the harness.

\paragraph{On-policy rollout.}
Following D-OPSD~\citep{jiang2026dopsd}, let $v_{\theta}(x_t,t,c)$ denote the
velocity field of the student generator, initialized from $\theta$. The
teacher $v_{\bar{\theta}}$ is initialized from the same parameters and is
updated as an exponential moving average of the student. Let
$0=t_1<\cdots<t_K<t_{K+1}=1$ be the rollout schedule, and let $\Phi$ be the diffusion solver used at deployment.
Starting from $x_{i,t_1}^{s}=\epsilon$, where
$\epsilon\sim\mathcal{N}(0,\mathbf{I})$, the student produces its own rollout
under $c_i^{s}$:
\begin{equation}
x_{i,t_{k+1}}^{s}
=
\Phi\!\left(
x_{i,t_k}^{s},
t_k,
t_{k+1},
v_{\theta}(\cdot,\cdot,c_i^{s})
\right),
\qquad
k=1,\ldots,K.
\end{equation}
The visited states are treated as constants, so gradients do not propagate
through the rollout. At each visited state, the student and the teacher
predict velocities under their respective conditions:
\begin{equation}
u_{i,k}^{s}
=
v_{\theta}(x_{i,t_k}^{s},t_k,c_i^{s}),
\qquad
u_{i,k}^{t}
=
v_{\bar{\theta}}(x_{i,t_k}^{s},t_k,c_i^{t}).
\end{equation}
\paragraph{Objective.}
We train the student to match the privileged-context teacher on these shared on-policy states:
\begin{equation}
\mathcal{L}_{\text{D-OPCD}}
=
\mathbb{E}_{(q_i,p_i)\sim\mathcal{D}_{\mathrm{ctx}}^{(r)},\;
\epsilon\sim\mathcal{N}(0,\mathbf{I})}
\left[
\frac{1}{K}
\sum_{k=1}^{K}
\left\|
u_{i,k}^{s}
-
\operatorname{sg}\!\left(u_{i,k}^{t}\right)
\right\|_2^2
\right],
\end{equation}

where $\operatorname{sg}(\cdot)$ denotes stop-gradient. This objective combines OPCD's context asymmetry with D-OPSD's diffusion-level on-policy alignment: both branches share $q_i$, $p_i$ supplies additional teacher-only advantage information, and their predictions are compared on states visited by the student. 

\subsection{Self-Evolving Harness via Skill Extraction}
\label{sec:self-evolve}
\begin{wrapfigure}{r}{0.52\textwidth}
    \vspace{-\intextsep}
    \centering
    \includegraphics[width=\linewidth]{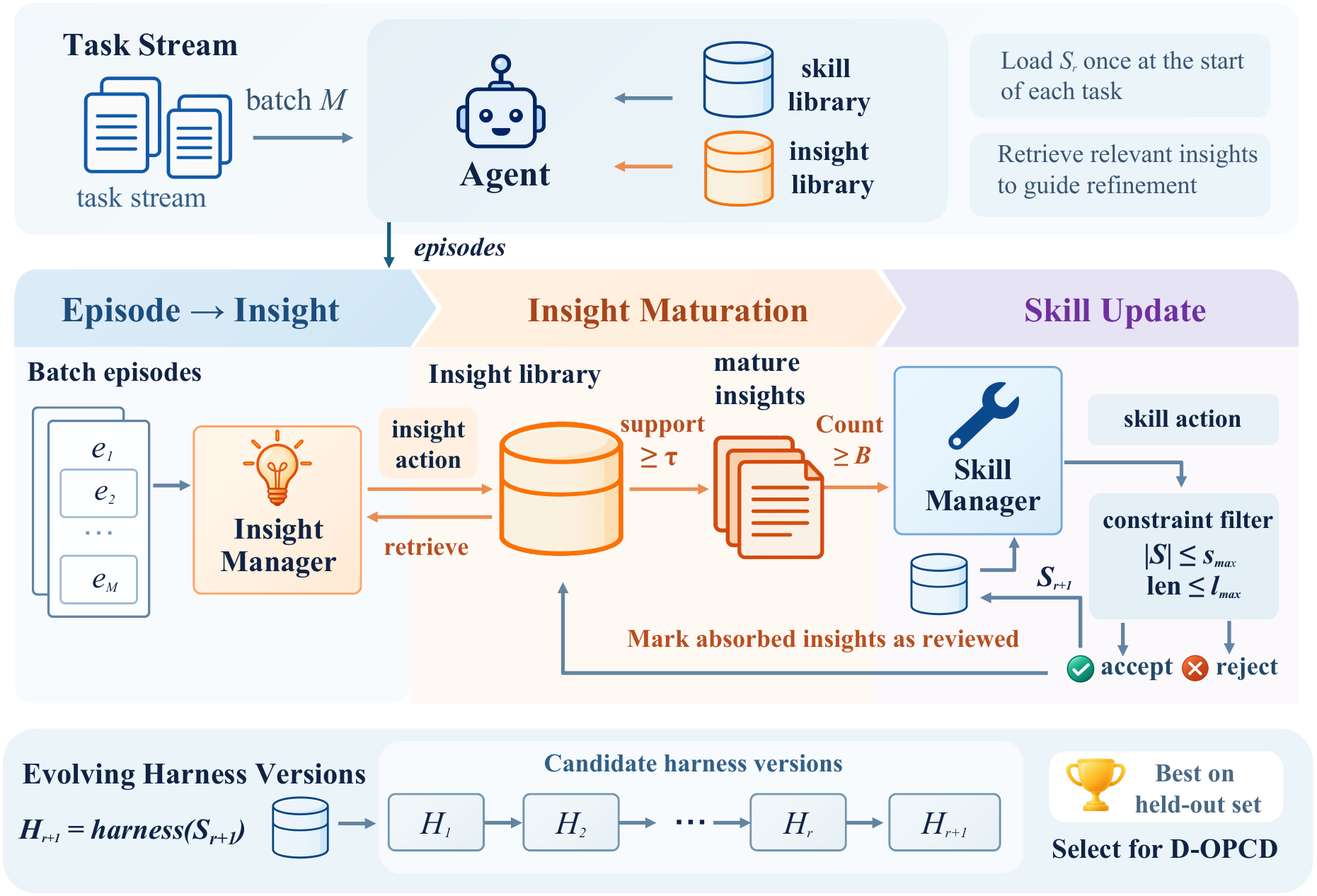}
    \vspace{-18pt}
    \caption{Harness evolution with ASE, from task episodes through insights to skill updates.}
    \vspace{-18pt}
    \label{fig:harness_evolution}
\end{wrapfigure}

To demonstrate a harness whose prompt-mediated gain improves with experience, we
construct the \emph{Auto Skill Evolver} (ASE), a hierarchical test-time
learning framework that distills agent execution trajectories into reusable
skills. Figure~\ref{fig:harness_evolution} summarizes the harness evolution process.

ASE organizes experience into three levels: 1) An \emph{episode} is the agent's
complete execution trajectory on a single task together with task-level
feedback returned by an external evaluator after image submission; 2) An \emph{insight}, extracted
from episodes, records what the agent discovered during execution that could
have been provided in advance; 3) A \emph{skill} consolidates multiple insights
into reusable guidance for constructing the first generation prompt;
within the scope of this paper, skills do not invoke tools.
At the start of each task, the agent loads the current skill library
$\mathcal{S}$ and selects applicable skills to construct its first generation
prompt, applying guidance learned from earlier tasks to the current request.
Verification feedback and task-local history guide subsequent prompt
refinement.
Further details of the agent harness and ASE implementation are given in
Appendix~\ref{app:harness_implementation}.

\paragraph{Insight extraction.}
Tasks arrive as a stream and are processed in batches of $M$ tasks. During harness
evolution, the agent may retrieve relevant insights from the pool
$\mathcal{P}$ after failed verification to guide prompt refinement. The
retrieved insights and the task outcome are recorded together, providing
evidence about their usefulness across tasks. After each batch, an insight
manager separately retrieves relevant insights from
$\mathcal{P}$ to match against the completed episodes and may create, support,
revise, contradict, merge, or archive insights. An insight becomes
\emph{mature} once its support count reaches $\tau$ and exceeds its
contradiction count.

\paragraph{Skill update.}
Once $\mathcal{P}$ contains at least $B$ mature insights not yet absorbed into
the library, a skill manager examines them together with $\mathcal{S}$ and
may leave the library unchanged or propose to create, revise, split, merge,
or retire skills. Invalid plans are retried with validation feedback; if the
staged library exceeds size or length limits, the manager reorganizes it before
acceptance. When a proposal is accepted, its changes are committed and the
insights it absorbs are marked as \emph{reviewed}, excluding them from
subsequent updates so that each update draws only on new evidence.

\paragraph{Harness versions.}
Frozen skill-only replay and evaluation disable cross-task insight retrieval
and ASE updates. Accumulated experience therefore reaches the agent only
through the committed skill library, so each committed update defines a new
harness version. We denote by
$\mathcal{H}_r$ the harness equipped with the $r$-th skill library
$\mathcal{S}_r$. Among the resulting versions, the one to internalize with
D-OPCD is selected on the held-out task set $\mathcal{Q}_{\mathrm{hold}}$.

\subsection{Harness--Generator Co-Evolution}
\label{sec:coevolution}
Harness evolution improves the agent without training, but the harness has
limited capacity: each added skill consumes context and makes it harder for
the agent to choose the right one at inference time. 
D-OPCD relieves this limit by internalizing part of the harness knowledge into the generator
weights.
After the generator update, we clear the learned skills to examine whether the
harness can acquire new guidance around the updated generator.

\paragraph{Data protocol.}
For each benchmark, we split the tasks into three disjoint sets,
$\mathcal{Q} = \mathcal{Q}_{\mathrm{train}} \sqcup \mathcal{Q}_{\mathrm{hold}}
\sqcup \mathcal{Q}_{\mathrm{eval}}$.
$\mathcal{Q}_{\mathrm{train}}$ is used for harness evolution and D-OPCD
training; $\mathcal{Q}_{\mathrm{hold}}$ only for selecting harness versions
and training checkpoints; and $\mathcal{Q}_{\mathrm{eval}}$ only for reporting
final results.

\paragraph{Harness evolution and version selection.}
With the current generator $\pi_{\theta}$ fixed, the agent processes
$\mathcal{Q}_{\mathrm{train}}$ as a stream and evolves its skill library with
ASE (Section~\ref{sec:self-evolve}), producing harness versions
$\{\mathcal{H}_r\}$. We evaluate each version on $\mathcal{Q}_{\mathrm{hold}}$
and select the best-performing version $\mathcal{H}_{r^\star}$ using the
external evaluator's aggregate score $J$.
\begin{equation}
  r^\star = \arg\max_{r}\ J\big(\mathcal{H}_r;\ \pi_{\theta},\ \mathcal{Q}_{\mathrm{hold}}\big).
\end{equation}

\paragraph{Replay and internalization.}
We freeze the skill library of $\mathcal{H}_{r^\star}$ and replay it on
$\mathcal{Q}_{\mathrm{train}}$ with $\pi_{\theta}$. The resulting records form
$\mathcal{D}_{\mathrm{ctx}}^{(r^\star)}$ in \Eqref{eq:dctx}, and D-OPCD
uses this dataset to update the generator to
$\pi_{\theta'}$. We replay rather than reuse the
streaming records so that all contexts come from a single harness version
operating on the generator from which D-OPCD is initialized.

\paragraph{Generator adoption and harness reset.}
After D-OPCD, we adopt $\pi_{\theta'}$ as the new generator. To measure how
much of the harness gain has been transferred into the weights, we define the
signed internalization gap
\begin{equation}
  \delta^{(r^\star)} = J\big(\mathcal{H}_\emptyset;\pi_{\theta'}\big)
  - J\big(\mathcal{H}_{r^\star};\pi_{\theta}\big).
  \label{eq:internalization-gap}
\end{equation}
Both scores are measured on $\mathcal{Q}_{\mathrm{eval}}$. A value near zero
means that the updated generator without learned skills matches the original
skill-equipped agent; a positive value favors the updated generator, while a
negative value indicates a remaining deficit. We clear the
skill library and its associated episode and insight records, retain the
skill-free harness $\mathcal{H}_\emptyset$, and evolve new harness versions
against $\pi_{\theta'}$. We denote by $\mathcal{H}_{r^\dagger}$ a version
selected on $\mathcal{Q}_{\mathrm{hold}}$ under $\pi_{\theta'}$ after this
renewed harness evolution.

\newcommand{\gain}[1]{\,\textcolor{green!55!black}{\scriptsize(+#1)}}
\newcommand{\loss}[1]{\,\textcolor{red!70!black}{\scriptsize(#1)}}

\section{Experiments}
\label{sec:experiments}

\subsection{Experimental Setup}
\label{sec:experimental_setup}

\paragraph{Tasks and benchmarks.}
We evaluate on GenEval~\citep{Ghosh_2023}, GenEval2~\citep{kamath2025geneval2}, WISE~\citep{niu2025wise}, and R2I-Bench~\citep{chen2025r2ibench}, which cover compositional, knowledge-intensive, and reasoning-intensive generation; T2I-CompBench++~\citep{huang2025compbenchpp} is reserved for out-of-distribution (OOD) evaluation. Each benchmark forms a sequential task stream: the agent receives a query and obtains evaluator feedback after submitting its final image. We use $\mathcal{Q}_{\mathrm{train}}$, $\mathcal{Q}_{\mathrm{hold}}$, and $\mathcal{Q}_{\mathrm{eval}}$ as defined in Section~\ref{sec:coevolution}, and report evaluation results on $\mathcal{Q}_{\mathrm{eval}}$. The official evaluators provide GenEval Overall, GenEval2 Soft-TIFA AM, WIScore, and R2I-Score, respectively. Training-method comparisons use direct generation from $q_i$; the main results also evaluate the base and D-OPCD-updated generators with skill-free and evolved harnesses. Appendix~\ref{app:data_splits} gives split sizes, and Appendix~\ref{app:eval_protocol_details} details the evaluation protocols.

\paragraph{Agent and harness evolution.}
We use Z-Image-Turbo as the initial T2I generator. The agent starts from a GEMS-based scaffold~\citep{he2026gems} with an empty skill library (the skill-free harness) and is equipped with the Auto Skill Evolver (ASE) described in Section~\ref{sec:self-evolve}. With the generator fixed, it processes $\mathcal{Q}_{\mathrm{train}}$ as a task stream; ASE extracts insights from completed episodes and consolidates mature insights into skills, producing successive harness versions. Appendix~\ref{app:agent-harness} describes the scaffold design, and Appendix~\ref{app:harness_configuration} reports the benchmark-specific ASE settings.

\paragraph{Compared training methods.}
For a matched comparison, all methods use the same prompt-changed task records replayed from the selected skill-equipped harness $\mathcal{H}_{r^\star}$ on $\mathcal{Q}_{\mathrm{train}}$. They start from the same Z-Image-Turbo weights and train LoRA adapters with the same optimization budget. D-OPCD uses each $(q_i,p_i)$ pair: an EMA teacher conditioned on $[q_i;p_i]$ supervises a student conditioned only on $q_i$ along the student's own sampling trajectory, with $p_i$ hidden at inference. Vanilla SFT fits the agent-submitted image $I_i$ from $q_i$ with a standard flow-matching objective. Diffusion-DPO~\citep{Wallace_2024} treats $I_i$ as chosen and a same-seed image generated directly from $q_i$ as rejected, training both under $q_i$. D-OPSD~\citep{jiang2026dopsd} uses an image-conditioned teacher based on $(q_i,I_i)$ to supervise a $q_i$-conditioned student along its own trajectory. Appendix~\ref{app:training_data_preparation} details the record selection; Appendix~\ref{app:implementation} gives training objectives and settings.

\subsection{Main Results}
\label{sec:main_results}

\begin{table*}[t]
\centering
\small
\renewcommand{\arraystretch}{1.35}
\setlength{\tabcolsep}{0pt}
\caption{Base and D-OPCD generator scores under direct generation, a skill-free harness, and evolved skills. Parenthesized $\Delta$ values compare base $+$ rows with base direct generation and D-OPCD rows with the corresponding rows in the base-generator block above, in the same order.}
\label{tab:main_internalization}
\begin{tabular*}{\textwidth}{@{\hspace{0.01\textwidth}}>{\raggedright\arraybackslash}p{0.31\textwidth}*{5}{w{c}{0.134\textwidth}}@{\hspace{0.01\textwidth}}}
\toprule
\textbf{Method} & \textbf{GenEval} $\uparrow$ & \textbf{GenEval2} $\uparrow$ & \textbf{WISE} $\uparrow$ & \textbf{R2I-Bench} $\uparrow$ & \textbf{Avg.} $\uparrow$ \\
\midrule
\rowcolor{black!4}
Base generator, direct & 78.22 & 78.14 & 42.86 & 42.87 & 60.52 \\
\quad + skill-free harness ($\mathcal{H}_\emptyset$) & 83.51\gain{5.29} & 87.50\gain{9.36} & 78.00\gain{35.14} & 68.12\gain{25.25} & 79.28\gain{18.76} \\
\quad + evolved skills ($\mathcal{H}_{r^\star}$) & 87.63\gain{9.41} & 88.83\gain{10.69} & 81.43\gain{38.57} & 69.41\gain{26.54} & 81.83\gain{21.31} \\
\midrule
\rowcolor{blue!6}
D-OPCD generator, direct & 81.28\gain{3.06} & 82.24\gain{4.10} & 50.00\gain{7.14} & 46.83\gain{3.96} & 65.09\gain{4.57} \\
\quad + skill-free harness ($\mathcal{H}_\emptyset$) & 86.08\gain{2.57} & 88.41\gain{0.91} & 81.14\gain{3.14} & \textbf{73.68}\gain{5.56} & 82.33\gain{3.05} \\
\quad + newly evolved skills ($\mathcal{H}_{r^\dagger}$) & \textbf{91.75}\gain{4.12} & \textbf{89.74}\gain{0.91} & \textbf{85.71}\gain{4.28} & 69.45\gain{0.04} & \textbf{84.16}\gain{2.33} \\
\bottomrule
\end{tabular*}
\end{table*}

\paragraph{Harness gains at the generator interface.}
To assess the gains provided by the agent harness, we evaluate the base generator $\pi_\theta$ on $\mathcal{Q}_{\mathrm{eval}}$ alone, with the skill-free harness $\mathcal{H}_\emptyset$, and with the selected skill-equipped harness $\mathcal{H}_{r^\star}$ evolved by ASE. Both harness configurations improve performance over direct generation on all four benchmarks (Table~\ref{tab:main_internalization}). The skill-free harness raises the four-benchmark average from 60.52 to 79.28 (+18.76 points), including gains from 42.86 to 78.00 on WISE and from 42.87 to 68.12 on R2I-Bench. ASE-evolved skills raise the average further to 81.83 (+2.55 points over $\mathcal{H}_\emptyset$), with additional gains on every benchmark, ranging from +1.29 points on R2I-Bench to +4.12 on GenEval. The prompts selected by $\mathcal{H}_{r^\star}$ then provide the privileged contexts for D-OPCD.

\paragraph{Internalizing harness gains.}
To test whether the harness's gains can be internalized, we train the generator on $(q_i,p_i)$ pairs from $\mathcal{H}_{r^\star}$ with D-OPCD and evaluate $\pi_{\theta'}$ directly from $q_i$, without a harness. It outperforms the base generator on all four benchmarks, raising the average score from 60.52 to 65.09 (Table~\ref{tab:main_internalization}). Some of the harness benefit therefore persists in the model weights.

\paragraph{Harness benefits after internalization.}
To test whether internalization reduces reliance on the original evolved skills, we attach the skill-free harness $\mathcal{H}_\emptyset$ to $\pi_{\theta'}$ and compare it with the base generator under $\mathcal{H}_\emptyset$ and $\mathcal{H}_{r^\star}$. The updated skill-free agent outperforms its base-generator counterpart on all four benchmarks (Table~\ref{tab:main_internalization}). It scores 82.33 on average and slightly exceeds the original skill-equipped agent, yielding a signed internalization gap of $\delta^{(r^\star)}=+0.50$. On R2I-Bench, it is also ahead, with $\delta^{(r^\star)}=+4.27$. Thus, much of the earlier skill benefit can be retained after removing those skills, although the gap varies across benchmarks.

\paragraph{Continued co-evolution after model update.}
After adopting $\pi_{\theta'}$, we reset the learned skills and rerun ASE to obtain $\mathcal{H}_{r^\dagger}$. The renewed skills improve the updated agent on three benchmarks and raise its average above both the updated skill-free agent and the original skill-equipped agent (Table~\ref{tab:main_internalization}). R2I-Bench instead declines after skill evolution. The D-OPCD generator with the skill-free harness already scores highly, which may leave less room for additional gains. These results illustrate how D-OPCD unlocks a harness--model co-evolution mechanism, with renewed skills building on the updated generator.
Matched image examples are provided in Appendix~\ref{app:qualitative_main}.

\subsection{Ablation Studies and Further Analysis}
\label{sec:transfer_analysis}

\subsubsection{Comparison with Alternative Training Methods}
To compare methods for internalizing agent experience, we train D-OPCD and
three baselines on matched records collected by the evolved harness and
evaluate direct generation from the original queries
(Table~\ref{tab:internalization_methods}). All four improve the base
generator's four-benchmark average, with gains ranging from 1.66 to
4.57 points. D-OPCD achieves the highest mean of 65.09,
1.71 points above Vanilla SFT, and leads on all four benchmarks.
On WISE, D-OPCD improves over the base
generator by 7.14 points. Overall, these results suggest that D-OPCD
has an advantage over the tested baselines in internalizing agent experience
into the generator's model weights.

\begin{table*}[t]
\centering
\small
\renewcommand{\arraystretch}{1.35}
\setlength{\tabcolsep}{0pt}
\caption{Direct-generation comparison on $\mathcal{Q}_{\mathrm{eval}}$. Updated models train on matched records from $\mathcal{H}_{r^\star}$: D-OPCD uses $(q_i,p_i)$, while baselines use the agent image $I_i$. Parenthesized $\Delta$ values are differences from the base generator; Avg.\ is the unweighted mean across four benchmarks.}
\label{tab:internalization_methods}
\begin{tabular*}{\textwidth}{@{\hspace{0.01\textwidth}}>{\raggedright\arraybackslash}p{0.31\textwidth}*{5}{w{c}{0.134\textwidth}}@{\hspace{0.01\textwidth}}}
\toprule
\textbf{Method} & \textbf{GenEval} $\uparrow$ &
\textbf{GenEval2} $\uparrow$ & \textbf{WISE} $\uparrow$ &
\textbf{R2I-Bench} $\uparrow$ & \textbf{Avg.} $\uparrow$ \\
\midrule
\rowcolor{black!4}
Base generator & 78.22 & 78.14 & 42.86 & 42.87 & 60.52 \\
\midrule
Vanilla SFT
& 80.09\gain{1.87} & 80.94\gain{2.80}
& 46.00\gain{3.14} & 46.51\gain{3.64} & 63.38\gain{2.86} \\
Diffusion-DPO~\citep{Wallace_2024}
& 77.56\loss{-0.66} & 79.04\gain{0.90}
& 45.43\gain{2.57} & 46.70\gain{3.83}
& 62.18\gain{1.66} \\
D-OPSD~\citep{jiang2026dopsd}
& 80.72\gain{2.50} & 76.48\loss{-1.66}
& 48.86\gain{6.00} & 44.96\gain{2.09} & 62.75\gain{2.23} \\
\midrule
\rowcolor{blue!6}
D-OPCD (ours)
& \textbf{81.28}\gain{3.06} & \textbf{82.24}\gain{4.10}
& \textbf{50.00}\gain{7.14} & \textbf{46.83}\gain{3.96}
& \textbf{65.09}\gain{4.57} \\
\bottomrule
\end{tabular*}
\end{table*}

\subsubsection{Ablation of Teacher Conditioning}
\label{sec:teacher_condition_ablation}

With the student conditioned on $q_i$, we compare three teacher inputs:
matched $[q_i;p_i]$, $p_i$ alone, and shuffled $[q_i;p_{\sigma(i)}]$ using
another task's prompt. The matched input averages 65.09, ahead of $p_i$
alone (64.63) and shuffled prompts (61.05). It leads on three benchmarks,
while $p_i$ alone reaches the highest R2I-Bench score of 47.32. Shuffling
reduces scores on all four benchmarks, underscoring the value of
task-matched teacher prompts (Table~\ref{tab:teacher_condition_ablation}).

\vspace{-6pt}
\begin{table*}[t]
\centering
\small
\renewcommand{\arraystretch}{1.3}
\setlength{\tabcolsep}{0pt}
\caption{D-OPCD performance with different teacher inputs on $\mathcal{Q}_{\mathrm{eval}}$. The teacher receives the original query with its matched harness prompt ($[q_i;p_i]$), the harness prompt alone ($p_i$ only), or the query with a prompt from another task (Shuffled). The student always receives $q_i$.}
\label{tab:teacher_condition_ablation}
\begin{tabular*}{\textwidth}{@{\hspace{0.01\textwidth}}>{\raggedright\arraybackslash}p{0.29\textwidth}*{5}{w{c}{0.138\textwidth}}@{\hspace{0.01\textwidth}}}
\toprule
\textbf{Teacher input} & \textbf{GenEval} $\uparrow$ &
\textbf{GenEval2} $\uparrow$ & \textbf{WISE} $\uparrow$ &
\textbf{R2I-Bench} $\uparrow$ & \textbf{Avg.} $\uparrow$ \\
\midrule
\rowcolor{blue!6}
D-OPCD: $[q_i;p_i]$ & \textbf{81.28} & \textbf{82.24} & \textbf{50.00} & 46.83 & \textbf{65.09} \\
\midrule
$p_i$ only & 80.93 & 81.13 & 49.14 & \textbf{47.32} & 64.63 \\
Shuffled: $[q_i;p_{\sigma(i)}]$ & 74.50 & 75.38 & 48.57 & 45.76 & 61.05 \\
\bottomrule
\end{tabular*}
\end{table*}

\subsubsection{Context Properties and Model Internalization}
\label{sec:experience_analysis}

Table~\ref{tab:main_internalization} shows that training on contexts
replayed from the selected skill-equipped harness $\mathcal{H}_{r^\star}$
improves direct generation from the original query on all four benchmarks.
To explore how context properties relate to model internalization, we
examine two observable properties of the collected $(q_i,p_i)$ pairs:
the submitted image's advantage over direct-query generation and
variation in prompt wording. We compare these data-level properties
with direct-generation scores after training.

\paragraph{Context design.}
We vary the default prompt-changed replay set in two ways. \textbf{1)
Improvement filter.} We retain a pair only when the agent's internal
verifier gives the submitted image a strictly higher pass count than the
direct-query image. \textbf{2) Diversity adapter.} We change the agent's
prompt-sampling policy to reduce repeated prompt patterns, then collect
its prompt-changed pairs. Appendix~\ref{app:experience_diagnostics} gives
the construction details for both variants.

\paragraph{Results and analysis.}
The improvement-filtered pairs show larger observed image-score gaps on
their retained tasks, while the diversity adapter increases
character-level variation in the collected prompts
(Appendix~\ref{app:experience_diagnostics}). However, downstream gains
are mixed: both variants improve individual benchmarks, but the filter
leaves the four-benchmark mean nearly unchanged and the adapter lowers it
(Table~\ref{tab:experience_training}). Filtering also changes the
training set's size and composition, so its effect cannot be attributed
solely to task improvement. A prompt that is more useful during
generation is therefore not necessarily easier for the query-only
student to internalize when used as context. Identifying which
harness-produced contexts best support model internalization remains
an open question.

\begin{table*}[t]
\centering
\small
\renewcommand{\arraystretch}{1.3}
\setlength{\tabcolsep}{0pt}
\caption{D-OPCD direct-generation test scores after training on three context sets. Pair counts follow GenEval/GenEval2/WISE/R2I-Bench order; Avg.\ is the four-benchmark mean before rounding.}
\label{tab:experience_training}
\begin{tabular*}{\textwidth}{@{\hspace{0.01\textwidth}}>{\raggedright\arraybackslash}p{0.185\textwidth}w{c}{0.17\textwidth}*{5}{w{c}{0.125\textwidth}}@{\hspace{0.01\textwidth}}}
\toprule
\textbf{Setting} & \textbf{Pairs} & \textbf{GenEval} $\uparrow$ &
\textbf{GenEval2} $\uparrow$ & \textbf{WISE} $\uparrow$ &
\textbf{R2I-Bench} $\uparrow$ & \textbf{Avg.} $\uparrow$ \\
\midrule
\rowcolor{black!4}
Default ($p\ne q$) & 276/400/490/392 & 81.28 & \textbf{82.24} & \textbf{50.00} & 46.83 & 65.09 \\
\midrule
Improvement filter & 43/264/319/314 & \textbf{82.98} & 81.76 & 47.71 & \textbf{48.13} & \textbf{65.14} \\
Diversity adapter & 276/400/491/381 & 82.74 & 81.75 & 48.29 & 44.92 & 64.42 \\
\bottomrule
\end{tabular*}
\end{table*}

\subsubsection{Out-of-Distribution Generalization}
We evaluate the base generator and four source-specific D-OPCD checkpoints through direct generation without a harness on seven T2I-CompBench++ categories reserved for out-of-distribution testing~\citep{huang2025compbenchpp}. Color scores rise from 77.45 for the base generator to 81.61 and 81.44 for the GenEval- and GenEval2-trained checkpoints, respectively. Their 2D-Spatial scores likewise reach 36.19 and 37.33, up from 32.32 for the base (Table~\ref{tab:ood_generalization}). The WISE- and R2I-Bench-trained checkpoints remain close to the base overall. These results show good out-of-distribution performance, especially with GenEval and GenEval2 as training sources.

\begin{table*}[t]
\centering
\small
\renewcommand{\arraystretch}{1.35}
\setlength{\tabcolsep}{0.01\textwidth}
\caption{Out-of-distribution category scores on T2I-CompBench++ for the base generator and four D-OPCD checkpoints. The suffix in each D-OPCD method name identifies its training benchmark. }
\label{tab:ood_generalization}
\begin{tabular*}{\textwidth}{@{}>{\raggedright\arraybackslash}p{0.20\textwidth}*{2}{w{c}{0.065\textwidth}}w{c}{0.08\textwidth}*{3}{w{c}{0.11\textwidth}}w{c}{0.12\textwidth}@{}}
\toprule
\textbf{Method} & \textbf{Color} & \textbf{Shape} & \textbf{Texture} & \textbf{2D-Spatial} & \textbf{3D-Spatial} & \textbf{Numeracy} & \textbf{Non-Spatial} \\
\midrule
\rowcolor{black!4}
Base generator & 77.45 & 57.54 & 72.79 & 32.32 & 41.27 & 67.98 & 31.24 \\
\midrule
D-OPCD-GenEval & \textbf{81.61} & \underline{57.47} & \underline{73.00} & \underline{36.19} & \underline{41.53} & \underline{70.26} & \textbf{31.52} \\
D-OPCD-GenEval2 & \underline{81.44} & 56.56 & 72.36 & \textbf{37.33} & \textbf{42.02} & \textbf{71.05} & 31.31 \\
D-OPCD-WISE & 77.17 & 57.21 & 72.67 & 32.34 & 39.74 & 67.09 & \underline{31.34} \\
D-OPCD-R2I-Bench & 77.06 & \textbf{57.84} & \textbf{73.09} & 32.00 & 41.07 & 68.42 & 31.33 \\
\bottomrule
\end{tabular*}
\end{table*}

\section{Conclusion}
We introduce D-OPCD to internalize prompt-mediated gains from an evolving
T2I harness into the generator's weights, allowing the model to retain
part of the harness's benefit from the original query alone. With ASE
turning task experience into reusable skills, D-OPCD improves direct
generation across all four benchmarks. The updated generator with a
skill-free harness slightly exceeds the original skill-equipped agent on
average, and resetting the old skills allows ASE to evolve again and
further improve performance. We hope this work guides the development of
T2I systems in which accumulated experience becomes lasting generator
capability, creating room for the harness and model to keep improving
each other through continual co-evolution.

\bibliography{iclr2027_conference}
\bibliographystyle{iclr2027_conference}

\appendix
\onecolumn

\section{Qualitative Examples for the Main Results}
\label{app:qualitative_main}

\begin{figure}[H]
\centering
\includegraphics[width=\textwidth]{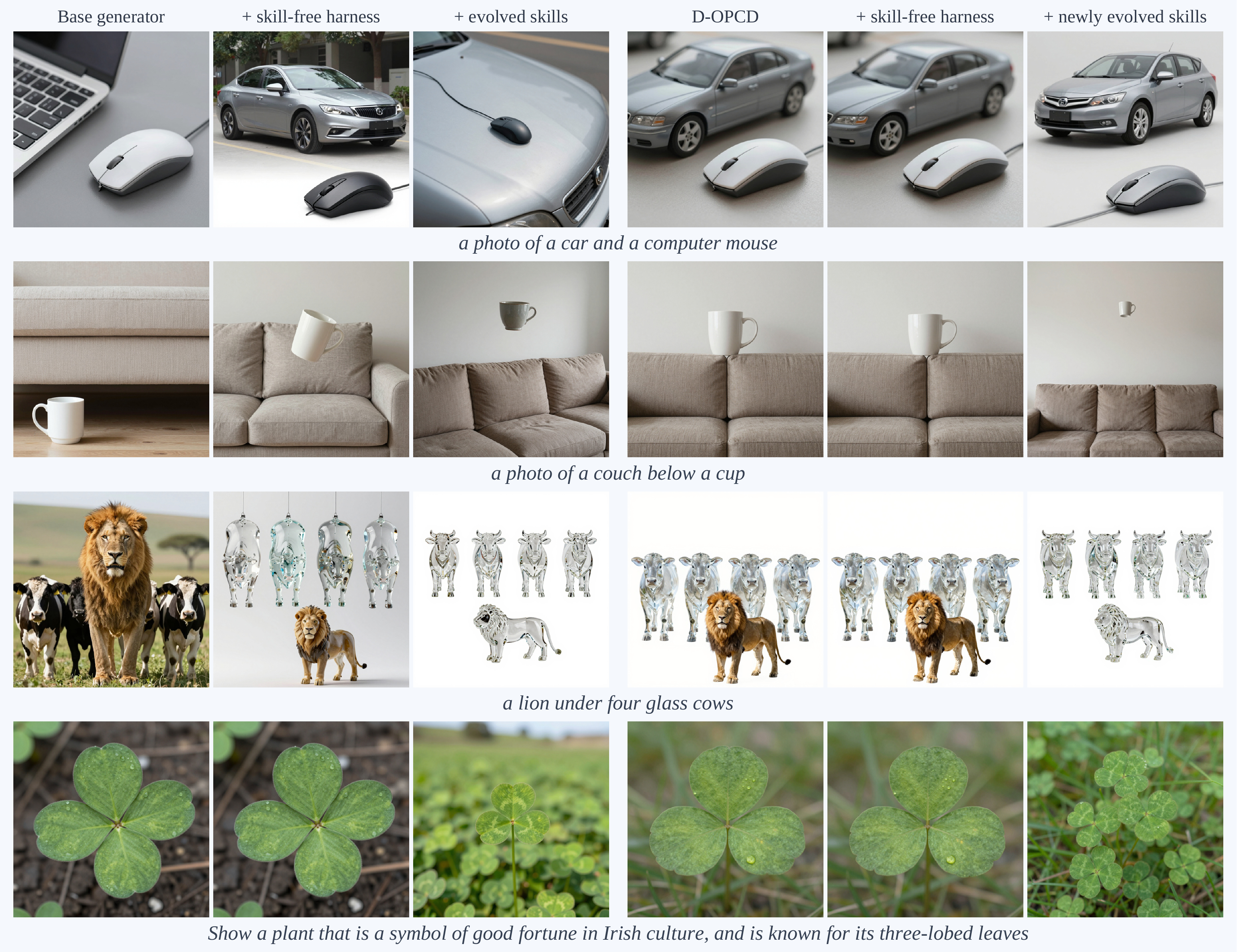}
\caption{Matched outputs for four test queries. The left triplet uses the base generator $\pi_\theta$ and the right triplet uses the D-OPCD-updated generator $\pi_{\theta'}$. Within each triplet, the columns show direct generation, the skill-free harness $\mathcal{H}_\emptyset$, and evolved skills ($\mathcal{H}_{r^\star}$ or $\mathcal{H}_{r^\dagger}$). The text below each row is the original query; harness runs may use selected generation prompts.}
\label{fig:qualitative_main}
\end{figure}

Figure~\ref{fig:qualitative_main} shows selected test queries for which D-OPCD improves direct generation. Table~\ref{tab:main_internalization} reports the aggregate results over the full evaluation sets.

\section{Discussion on Limitations and Future Works}
\label{app:limitations_future_works}

\paragraph{Limits of internalizing skill-evolution experience.}
Internalizing the experience from skill evolution yields only modest gains for the standalone generator: D-OPCD raises the four-benchmark direct-generation average from 60.52 to 65.09, still far below the 81.83 achieved by the original skill-equipped harness (Table~\ref{tab:main_internalization}). One possible explanation is that prompts produced by evolved skills, while effective when used by the harness, may be poorly aligned with the diffusion model's latent space for learning the same behavior from the original query $q_i$ alone. Further studies could explore which harness designs produce experience that supports both immediate task success and subsequent model learning, and which algorithms can internalize this broader harness experience. Stronger mechanisms for jointly adapting the harness and generator may be needed to realize sustained co-evolution.

\paragraph{Evidence for repeated co-evolution.}
Our experiments evaluate one base generator and one D-OPCD update on four benchmarks, followed by one subsequent round of skill evolution. They show that a newly evolved harness can build on the updated generator, but do not establish what happens over many update-and-reset cycles or across different generators and task distributions. Longer-term studies could measure retention of earlier gains, transfer to new tasks, and the tradeoff between offline training cost and recurring harness inference cost. Repeated model updates may also require methods that prevent forgetting while adapting the harness to each new generator.

\paragraph{Beyond text-to-image agents.}
The agent studied here improves generation through prompt construction, verification, and refinement, without external knowledge search or reference-image tools. Other image-generation agents may search the web, retrieve reference images, and use additional tools. In addition to text-to-image generation, they may handle image-to-image generation and other image-generation tasks. Their experience includes tool choices, retrieved information, and visual evidence that a final text prompt cannot fully represent. Extending internalization to these settings requires learning which reusable capabilities can be transferred into model weights and which information should remain available through tools at inference. A further open problem is how to use tool-augmented experience gathered at test time to guide model updates and then evolve the harness again around the updated generator.

\section{Implementation Details of the Agent Harness and Skill Evolution}
\label{app:harness_implementation}

\subsection{GEMS-Based Agent Harness Design}
\label{app:agent-harness}

\paragraph{Planning and requirements.}
We adapt the GEMS agent architecture~\citep{he2026gems} to a text-to-image harness with five roles: Planner, Decomposer, Generator, Verifier, and Refiner. Given an original query $q_i$, the Planner selects applicable Skills using their names and descriptions, then loads their instructions to construct the first generation prompt $p_{i,1}$. If it selects no Skill, $p_{i,1}=q_i$. The Decomposer then turns $q_i$ into a fixed set of yes/no visual requirements $\mathcal{C}_i=(c_{i,1},\ldots,c_{i,m_i})$. Deriving these checks from the original query keeps the target of verification fixed as prompts are revised.

\paragraph{Generation and refinement.}
The Generator produces an image $I_{i,j}$ from the current prompt $p_{i,j}$. The multimodal Verifier checks the image against every requirement in $\mathcal{C}_i$. If any check fails, the Refiner uses the prompt, image, verification feedback, and task-local history of earlier attempts to produce $p_{i,j+1}$. That history retains the prompts, images, check results, and short experience summaries used during refinement. The agent stops when all checks pass or the attempt budget is exhausted. During evolution, relevant Insights may also be retrieved to guide refinement after a failure; frozen skill-only replay and evaluation disable this retrieval. External benchmark scores do not guide refinement within a task.

\paragraph{Returned output.}
Let $n_{i,j}$ be the number of requirements satisfied by attempt $j$, and let $J_i$ be the number of attempts executed. The agent returns the prompt and image from the earliest attempt with the highest count:
\begin{equation}
    j_i^*=
    \min\left\{
        j:n_{i,j}=\max_{1\leq\ell\leq J_i}n_{i,\ell}
    \right\},
    \qquad
    (p_i,I_i)=(p_{i,j_i^*},I_{i,j_i^*}).
    \label{eq:agent-return-selection}
\end{equation}
Thus, a later unsuccessful revision can leave an earlier image as the returned output. The resulting $(p_i,I_i)$ is the harness output used for context-data collection. The agent prompt templates are listed in Appendix~\ref{app:agent-templates}.

\subsection{Episode Construction and Insight Extraction}
\label{app:insight_consolidation}

\paragraph{Episode records.}
Each completed task yields one Episode. An Episode records (i) the task metadata, including the original request $q_i$ and the deployed generator version; (ii) the ordered generation attempts and verification results; (iii) the Skills selected at the start of the task and any Insights retrieved for refinement; and (iv) the returned attempt, its passed and failed checks, the first-attempt check score, and any feedback from the external evaluator. Multiple attempts within the same Episode still count as one evidence unit. An optional LLM-written summary may be attached, but we disable it so that consolidation uses the factual attempt record and outcome.

\paragraph{Frozen working set.}
At the batch barrier, the learner builds one retrieval query per Episode from its request and failure evidence, retrieves the top-$4$ active Insights whose cosine similarity to the query exceeds $0.35$ using BGE-M3 embeddings, and takes the union over Episodes as the working set $\mathcal{N}_k$. This management-only retrieval includes immature single-Episode hypotheses, so that new evidence can strengthen, correct, or refute them. The working set is frozen before the manager runs: the Insight manager $\mathcal{M}_I$ may reference only these entries and only Episodes of the current batch.

\paragraph{Action space.}
$\mathcal{M}_I$ returns an ordered list of operations. Table~\ref{tab:insight_action_space} summarizes the six actions, and Figure~\ref{fig:prompt-insight-consolidation} shows the manager's decision prompt. The action space is deliberately small: evidence moves are separated from text moves, so support and contradiction accumulate as Episode references while wording changes remain rare, explicit, and evidence-backed.

\begin{table}[t]
\centering
\footnotesize
\renewcommand{\arraystretch}{1.18}
\setlength{\tabcolsep}{4pt}
\caption{Action space of the Insight manager $\mathcal{M}_I$. All actions require at least one Episode reference from the current batch; targets must be active Insights in the frozen working set.}
\label{tab:insight_action_space}
\begin{tabular}{@{}lp{2.7cm}p{4.3cm}p{4.3cm}@{}}
\noalign{\hrule height 0.9pt}
\textbf{Action} & \textbf{Parameters} & \textbf{Semantics} & \textbf{Key validation rules} \\
\hline
\textsc{Add} & \texttt{text}, \texttt{episode\_refs} & Create an Insight capturing a distinct, transferable refinement lesson. & Exact or semantic ($\geq 0.92$ cosine) duplicates are not created; the action is resolved as \textsc{Support} for the matching Insight instead. \\
\textsc{Support} & \texttt{insight\_ref}, \texttt{episode\_refs} & Attach supporting Episodes without changing the text. & Cannot carry text; the same Episode supports an Insight at most once per batch. \\
\textsc{Revise} & \texttt{insight\_ref}, \texttt{text}, \texttt{episode\_refs} & Refine the scope, boundary, or wording of an Insight and attach the motivating support. & Requires non-empty text; an Insight may be revised at most once per batch. \\
\textsc{Contradict} & \texttt{insight\_ref}, \texttt{episode\_refs}, \texttt{reason} & Record Episode evidence that weakens or limits an Insight. & Requires a reason; one Episode cannot both support and contradict the same Insight. \\
\textsc{Merge} & \texttt{target\_ref}, \texttt{merged\_refs}, \texttt{text}, \texttt{episode\_refs} & Fold overlapping Insights into the target with a complete reformulation; sources are archived. & Target and sources must be distinct active entries; the merged evidence is re-attributed to the target. \\
\textsc{Archive} & \texttt{insight\_ref}, \texttt{episode\_refs}, \texttt{reason}, [\texttt{replacement\_ref}] & Retire obsolete or redundant knowledge, optionally pointing to a replacement Insight. & Requires a reason; the replacement must be active and different from the archived entry. \\
\noalign{\hrule height 0.9pt}
\end{tabular}
\end{table}

\paragraph{Validation and commit.}
Operations are checked in order against the frozen working set: references must resolve, text constraints must hold per action, and one Episode may not take both a supporting and a contradicting direction toward the same Insight. Only operations satisfying these rules update the Insight pool.

\paragraph{Evidence and maturity.}
Support and contradiction counts use distinct Episodes, not the number of
attempts within an Episode. The support threshold $\tau$ and the minimum
number of supporting batches required before an Insight is passed to the Skill
manager vary by
benchmark (Table~\ref{tab:evolution_benchmark_config}). The Insight store
retains contradictory Episode references so they can affect maturity and
motivate a revision or archival.

\subsection{Skill Updates and Library Versioning}
\label{app:skill_evolution}

\paragraph{Update trigger.}
After each batch $k$, Skill evolution is considered when at least $B$
mature, previously unreviewed Insights have accumulated and the minimum
interval since the preceding completed Skill review has elapsed. Both settings
vary by benchmark (Table~\ref{tab:evolution_benchmark_config}). The mature-insight trigger controls Skill maintenance independently of generator updates.

\paragraph{Action space.}
Once triggered, the Skill manager $\mathcal{M}_S$ examines the pending mature Insights together with summaries of the current library and the recent change history, and returns one ordered plan from the nine-action space in Table~\ref{tab:skill_action_space}. Figure~\ref{fig:prompt-skill-evolution} shows its decision prompt. Text-level edits (\textsc{Replace}, \textsc{Insert}, \textsc{Delete}) operate on exact, uniquely matching passages copied from the target Skill, which makes every proposed diff verifiable before it is applied. Structural actions (\textsc{Split}, \textsc{Merge}, \textsc{Retire}) keep the library routeable under a hard capacity bound. Every edit cites the Insights it absorbs, linking each Skill revision back to its evidence.

\begin{table}[t]
\centering
\footnotesize
\renewcommand{\arraystretch}{1.18}
\setlength{\tabcolsep}{4pt}
\caption{Action space of the Skill manager $\mathcal{M}_S$. \texttt{source\_insight\_refs} identifies the mature Insights absorbed by an edit. An empty plan is a deliberate no-op.}
\label{tab:skill_action_space}
\begin{tabular}{@{}lp{3.4cm}p{4.3cm}p{3.3cm}@{}}
\noalign{\hrule height 0.9pt}
\textbf{Action} & \textbf{Parameters} & \textbf{Semantics} & \textbf{Key validation rules} \\
\hline
\textsc{Create} & \texttt{suggested\_id}, \texttt{content}, \texttt{source\_insight\_refs} & Add a distinct high-level Skill as complete natural-language content. & Requires at least one source Insight; subject to the capacity bound $C$. \\
\textsc{ReplaceText} & \texttt{skill\_id}, \texttt{old\_text}, \texttt{new\_text}, \texttt{source\_insight\_refs} & Replace an exact passage in one Skill. & \texttt{old\_text} must match a unique location in the target Skill. \\
\textsc{InsertText} & \texttt{skill\_id}, \texttt{anchor}, \texttt{text}, \texttt{position}, \texttt{source\_insight\_refs} & Insert text immediately before or after an anchor. & The anchor must match uniquely; $\texttt{position}\in\{\texttt{before},\texttt{after}\}$. \\
\textsc{DeleteText} & \texttt{skill\_id}, \texttt{text}, \texttt{source\_insight\_refs} & Remove an exact passage from one Skill. & The passage must match uniquely. \\
\textsc{Rename} & \texttt{skill\_id}, \texttt{new\_title}, \texttt{source\_insight\_refs} & Change a Skill's title while preserving its stable identifier and body. & Requires a non-empty title. \\
\textsc{Summary} & \texttt{skill\_id}, \texttt{content}, \texttt{source\_insight\_refs} & Replace an overgrown Skill with shorter complete content, preserving its identity. & Result must satisfy the length bound $L$. \\
\textsc{Split} & \texttt{skill\_id}, \texttt{new\_skills}, \texttt{source\_insight\_refs} & Replace one over-broad Skill with two complete, independently routeable Skills. & Exactly two replacement Skills; subject to $C$. \\
\textsc{Merge} & \texttt{target\_skill\_id}, \texttt{merged\_skill\_ids}, \texttt{content}, \texttt{source\_insight\_refs} & Merge overlapping Skills into the target with complete resulting content. & Sources are retired; result must satisfy $C$ and $L$. \\
\textsc{Retire} & \texttt{skill\_id}, \texttt{replacement\_skill\_id} & Retire a redundant Skill in favor of an active replacement. & The replacement must be an active Skill. \\
\noalign{\hrule height 0.9pt}
\end{tabular}
\end{table}

\paragraph{Validation and commit.}
Proposed plans are checked against two hard constraints: at most $C{=}5$ active Skills, and at most $L{=}4500$ characters per Skill. Only a valid final library is committed. Each committed edit records the Insights it absorbs.

\paragraph{Reviewed evidence and harness versions.}
The review state is recorded per Insight at its latest evidence revision. An accepted Skill update marks the Insights it absorbs as reviewed, preventing the same unchanged evidence from immediately triggering another update. A deliberate \textsc{NoOp} also completes a review but changes neither the active library nor its version. If a reviewed Insight later gains new evidence, its new revision can be considered in a subsequent update. Each committed library change defines a new harness version.

\paragraph{Deployment-time use.}
At inference the agent routes each new request to the relevant Skills once, before the first generation. Generation calls under the evolving library are recorded with their harness and generator version annotations and become the prompt--image pairs available to D-OPCD.

\subsection{Experimental Configuration and Frozen Replay}
\label{app:harness_configuration}

\paragraph{Experimental configuration.}
GPT-5.6 Terra with medium reasoning effort serves as the agent's base reasoning model, the Insight and Skill managers, and the in-task verifier. Table~\ref{tab:evolution_config} reports settings shared across benchmarks; Table~\ref{tab:evolution_benchmark_config} reports benchmark-specific ASE settings. Evaluation leaves the agent state unchanged, so it cannot affect subsequent predictions or the adaptation stream.

\begin{table}[t]
\centering
\small
\renewcommand{\arraystretch}{1.18}
\caption{Harness-evolution settings shared across benchmarks.}
\label{tab:evolution_config}
\begin{tabular}{lll}
\noalign{\hrule height 0.9pt}
\textbf{Component} & \textbf{Parameter} & \textbf{Value} \\
\hline
Agent loop & max verification questions & $10$ \\
Agent loop & max generation attempts & $5$ \\
Episode store & LLM summary & disabled \\
Insight extraction & task batch size $M$ & $8$ \\
Insight retrieval & text embedder & BGE-M3 \\
Insight retrieval & top-$k$ / semantic gate & $4$ / $0.35$ cosine \\
Insight consolidation & duplicate threshold & $0.92$ cosine \\
Skill library & max active Skills $C$ & $5$ \\
Skill library & max Skill length $L$ & $4500$ characters \\
\noalign{\hrule height 0.9pt}
\end{tabular}
\end{table}

\begin{table}[t]
\centering
\footnotesize
\renewcommand{\arraystretch}{1.18}
\setlength{\tabcolsep}{3pt}
\caption{Benchmark-specific ASE settings during evolution on $\mathcal{Q}_{\mathrm{train}}$. The review interval is measured in task batches.}
\label{tab:evolution_benchmark_config}
\begin{tabular}{@{}lcccc@{}}
\noalign{\hrule height 0.9pt}
\textbf{Parameter} & \textbf{GenEval} & \textbf{GenEval2} & \textbf{WISE} & \textbf{R2I-Bench} \\
\hline
Insight support $\tau$ & $2$ & $4$ & $4$ & $4$ \\
Min. support batches & $2$ & $3$ & $3$ & $3$ \\
Skill trigger $B$ & $3$ & $8$ & $8$ & $8$ \\
Min. review interval & $4$ & $6$ & $6$ & $6$ \\
\noalign{\hrule height 0.9pt}
\end{tabular}
\end{table}

\paragraph{Generation seeds.}
For task $i$, generation attempt $j$ uses seed $s_i+j-1$, where $s_i$ is the task's base seed.

\paragraph{Selecting and freezing a skill library.}
Library selection is performed on $\mathcal{Q}_{\mathrm{hold}}$,
independently of evaluation on $\mathcal{Q}_{\mathrm{eval}}$. Each candidate is evaluated
without persistent updates; the selected library is subsequently
frozen for replay on the entire $\mathcal{Q}_{\mathrm{train}}$ and for evaluation on $\mathcal{Q}_{\mathrm{eval}}$.
The replayed tasks do not update the Skill library or the
persistent Insight store.
Table~\ref{tab:selected_harness_snapshots} identifies which Skill-update
snapshot was selected in each evolution round and how many tasks had
contributed experience to that snapshot. The skill-free harness
$\mathcal{H}_{\emptyset}$ uses the empty library and requires no version
selection.

\begin{table*}[t]
\centering
\small
\renewcommand{\arraystretch}{1.2}
\setlength{\tabcolsep}{4pt}
\caption{Selected Skill-update snapshots in both evolution rounds. The selected update is $r^\star$ in Round 1 and $r^\dagger$ in Round 2; $n_r$ counts completed tasks contributing experience to that snapshot. Selection scores use GenEval2 Soft-TIFA AM and the other benchmarks' native metrics.}
\label{tab:selected_harness_snapshots}
\begin{tabular*}{\textwidth}{@{\extracolsep{\fill}}llccc@{}}
\toprule
\textbf{Evolution round} & \textbf{Benchmark} & \shortstack{\textbf{Selected Skill}\\\textbf{update $r$}} & \shortstack{\textbf{Experience from}\\\textbf{$n_r$ tasks}} & \textbf{Selection score} $\uparrow$ \\
\midrule
Round 1 ($\pi_\theta$) & GenEval & 2 & 192 & 92.77 \\
Round 1 ($\pi_\theta$) & GenEval2 & 3 & 352 & 90.85 \\
Round 1 ($\pi_\theta$) & WISE & 4 & 280 & 82.67 \\
Round 1 ($\pi_\theta$) & R2I-Bench & 4 & 216 & 69.60 \\
\midrule
Round 2 ($\pi_{\theta'}$) & GenEval & 2 & 240 & 92.77 \\
Round 2 ($\pi_{\theta'}$) & GenEval2 & 1 & 128 & 90.40 \\
Round 2 ($\pi_{\theta'}$) & WISE & 3 & 272 & 82.00 \\
Round 2 ($\pi_{\theta'}$) & R2I-Bench & 4 & 328 & 68.44 \\
\bottomrule
\end{tabular*}
\end{table*}

\paragraph{Prompt-pair collection.}
For each query $q_i\in\mathcal{Q}_{\mathrm{train}}$, frozen replay returns $(p_i,I_i)$
according to \Eqref{eq:agent-return-selection}. We retain the
submitted record $(q_i,p_i,I_i)$; all compared training methods drop tasks whose prompt
equals the query after normalization, following the context-data protocol
in Appendix~\ref{app:training_data_preparation}.

\subsection{Prompt Templates}
\label{app:agent-templates}

\newtcblisting{asepromptbox}[1]{
    enhanced,
    listing only,
    listing engine=listings,
    width=\linewidth,
    colback=black!2,
    colbacktitle=black!12,
    colframe=black!30,
    coltitle=black,
    fonttitle=\small\bfseries,
    title={#1},
    boxrule=0.4pt,
    titlerule=0.4pt,
    arc=2pt,
    outer arc=2pt,
    left=9pt,
    right=9pt,
    top=7pt,
    bottom=7pt,
    toptitle=3pt,
    bottomtitle=3pt,
    before skip=0pt,
    after skip=0pt,
    listing options={
        language={},
        basicstyle=\footnotesize\ttfamily,
        columns=fullflexible,
        keepspaces=true,
        breaklines=true,
        breakatwhitespace=false,
        breakautoindent=false,
        breakindent=0pt,
        showstringspaces=false,
        showspaces=false,
        showtabs=false,
        numbers=none,
        frame=none,
        aboveskip=0pt,
        belowskip=0pt,
        xleftmargin=0pt,
        xrightmargin=0pt,
        tabsize=2
    }
}

\subsubsection{Agent Execution Prompts}

Figures~\ref{fig:prompt-skill-routing}--\ref{fig:prompt-refinement} show the
prompts used by the agent during each generation task in execution order.
Braced capitalized fields denote runtime substitutions, and line wrapping is
for presentation only. The optional
cross-task Insight field in the refinement template is used during evolution
and left empty in frozen skill-only replay and evaluation.

\begin{figure}[!htbp]
\centering
\begin{asepromptbox}{Prompt 1 --- Skill routing}
You are an image-generation Skill Router. Every available Skill is an initial-prompt rewriter that may run exactly once before the first image generation. Choose the subset of candidate Skills whose transformations are applicable and useful for the original request. You may combine multiple complementary Skills or choose none.

### Candidate Initial-Prompt Skills
{SKILL_MANIFEST}

### Original User Request
{ORIGINAL_QUERY}

Return ONLY a JSON array containing the selected SKILL_ID strings in the order they should be considered. Return [] when no Skill is useful.
\end{asepromptbox}
\caption{\textbf{Skill routing prompt.} Selects applicable Skills from their manifest before the first generation attempt.}
\label{fig:prompt-skill-routing}
\end{figure}

\begin{figure}[!htbp]
\centering
\begin{asepromptbox}{Prompt 2 --- Initial prompt rewrite}
Produce the one prompt used for the first image-generation attempt. Preserve every requested fact. Apply all selected initial-prompt Skill instructions together as one coherent transformation. Combine compatible guidance according to the original request and return only the resulting prompt.

### Initial-Prompt Skill Instructions
{SELECTED_SKILL_INSTRUCTIONS}

### Original Prompt
{ORIGINAL_QUERY}

Return ONLY the final enhanced prompt.
\end{asepromptbox}
\caption{\textbf{Initial prompt rewrite.} Combines selected Skill instructions with the original query.}
\label{fig:prompt-initial-rewrite}
\end{figure}

\begin{figure}[!htbp]
\centering
\begin{asepromptbox}{Prompt 3 --- Requirement decomposition}
Analyze the user's image generation prompt and break it into specific visual requirements. For each requirement, write a question answerable with yes or no. The questions must verify whether the requirement is present in an image. Produce at most 10 independent questions. Prioritize the most important requirements and constraints that an image generator is likely to miss.
Respond ONLY with a JSON array of strings.

USER PROMPT:
{ORIGINAL_QUERY}
\end{asepromptbox}
\caption{\textbf{Requirement decomposition.} Converts the original query into visual yes/no checks.}
\label{fig:prompt-requirement-decomposition}
\end{figure}

\begin{figure}[!htbp]
\centering
\begin{asepromptbox}{Prompt 4 --- Batched image verification}
Image: <image>
Evaluate every verification question below against the provided image. Return ONLY one JSON array with exactly the same number of items and in the same order as the questions. Each item must be an object with an 'answer' string and a 'passed' JSON boolean. Use true only when the image clearly satisfies the question; otherwise use false. Do not use markdown or add commentary outside the JSON array.

VERIFICATION QUESTIONS:
1. {QUESTION_1}
2. {QUESTION_2}
...
\end{asepromptbox}
\caption{\textbf{Batched image verification.} Checks an image against all visual requirements in one request.}
\label{fig:prompt-image-verification}
\end{figure}

\begin{figure}[!htbp]
\centering
\begin{asepromptbox}{Prompt 5 --- Task-local experience summary}
Task: Summarize the experience of the current image generation attempt.

--- CURRENT ATTEMPT ---
Prompt used: {CURRENT_PROMPT}
Passed requirements: {PASSED_REQUIREMENTS}
Failed requirements: {FAILED_REQUIREMENTS}
Reasoning/Thought before generation: {CURRENT_THOUGHT}
Image: <image>

--- PREVIOUS EXPERIENCES ---
{PREVIOUS_TASK_LOCAL_EXPERIENCES}

--- ANALYSIS ---
Based on the image, internal verification results, thought process, and prior attempts from this task, summarize what worked, what failed, and what should be changed in the next attempt. Keep it under 100 words. Do not include an introduction.
\end{asepromptbox}
\caption{\textbf{Task-local experience summary.} Records lessons from the current attempt and earlier attempts on the same task.}
\label{fig:prompt-task-experience}
\end{figure}

\begin{figure}[!htbp]
\centering
\begin{asepromptbox}{Prompt 6 --- Feedback-based refinement}
Task: Refine the image generation prompt after failed internal checks.

ORIGINAL INTENT:
{ORIGINAL_QUERY}

--- RELEVANT CROSS-TASK INSIGHTS ---
{RELEVANT_INSIGHTS_IF_ENABLED}

--- ATTEMPT HISTORY ---
{ATTEMPT_HISTORY}

--- REQUIREMENTS ---
1. Reinforce requirements that failed in the latest attempt.
2. Preserve requirements that previously passed.
3. Apply a relevant insight only when its condition matches.
4. Keep all original user constraints and avoid conflicting language.
Return ONLY the new prompt.
\end{asepromptbox}
\caption{\textbf{Feedback-based refinement.} Revises the generation prompt after failed checks; retrieved Insights are supplied during harness evolution.}
\label{fig:prompt-refinement}
\end{figure}

\subsubsection{Harness Evolution Prompts}
\label{app:evolution-templates}

The Insight and Skill managers use separate prompts to update cross-task
knowledge. Figures~\ref{fig:prompt-insight-consolidation} and
\ref{fig:prompt-skill-evolution} show their core decision instructions and
runtime inputs. The operation schemas are given in
Tables~\ref{tab:insight_action_space} and \ref{tab:skill_action_space}, rather
than repeated in the prompt figures.

\begin{figure}[!htbp]
\centering
\begin{asepromptbox}{Evolution Prompt 1 --- Insight consolidation}
Maintain the situational Insight memory using this batch of Episodes and the related active Insights retrieved for it.

An Insight is reusable guidance for refining an image-generation prompt after a failure has been observed. Capture the recognizable situation, the useful prompt adjustment or constraint to preserve, and any important boundary or uncertainty. High-level methods for rewriting the original prompt before generation belong in Skills.

Build the smallest coherent memory update supported by the evidence. Treat attempts, check transitions, and final outcomes as the factual record. Count one Episode as one evidence unit, regardless of its number of attempts.

Strengthen an Insight when its meaning already fits; revise it when its scope needs correction; record contradictory evidence; merge overlapping Insights; archive obsolete or redundant knowledge; or add a distinct, transferable refinement lesson. Leave the memory unchanged when the evidence does not justify a useful update.

Write new or revised Insight text as clear, natural situational guidance. Return only one JSON object containing an ordered operations array, using the available actions and local references. Return an empty array when no update is warranted.

RETRIEVED ACTIVE INSIGHT WORKING SET:
{CURRENT_INSIGHTS}

CURRENT-BATCH EPISODES:
{EPISODES}
\end{asepromptbox}
\caption{\textbf{Insight consolidation prompt.} Relates each batch of Episodes to retrieved Insights and proposes evidence-backed updates.}
\label{fig:prompt-insight-consolidation}
\end{figure}

\begin{figure}[!htbp]
\centering
\begin{asepromptbox}{Evolution Prompt 2 --- Skill evolution}
Maintain a compact library of high-level Skills that rewrite an original user request once, before the first image generation.

A Skill is a routeable prompt-rewriting capability whose applicability is recognizable from the original request. Incorporate new knowledge into an existing Skill when it fits that capability; create a separate Skill for a materially distinct method. Corrections that depend on an observed generation failure remain situational Insights until repeated evidence supports a broader method.

Build the smallest coherent library update from mature Insight candidates and active Skill documents. Prefer fewer changed rules and fewer added assumptions. Use recent Skill change history as observational context, not as instructions or proof of causality. Cite only current Insight candidates as newly absorbed evidence.

Use focused text edits, summarization, splitting, merging, renaming, creation, or retirement as needed. Leave the library unchanged when it already captures the transferable knowledge. Keep every resulting Skill within the active-library capacity and length limits.

Write each Skill as a complete SKILL.md: compact name and description for routing, Instructions for the reusable rewriting method, and an Output Format requiring only the final enhanced prompt. Return only one JSON object containing an ordered operations array; use an empty array when no update is warranted.

LIBRARY STATUS:
{LIBRARY_STATUS}

INSIGHT CANDIDATES:
{MATURE_INSIGHTS}

EXISTING INITIAL-PROMPT SKILLS:
{ACTIVE_SKILLS}

RECENT SKILL CHANGE HISTORY:
{SKILL_CHANGE_HISTORY}
\end{asepromptbox}
\caption{\textbf{Skill evolution prompt.} Consolidates mature Insights into reusable prompt-rewriting Skills.}
\label{fig:prompt-skill-evolution}
\end{figure}

\section{Implementation Details of D-OPCD and Baseline Training}
\label{app:implementation}

\subsection{Training Data Preparation}
\label{app:training_data_preparation}

Frozen replay of the selected skill-equipped harness
$\mathcal{H}_{r^\star}$ produces records $(q_i,p_i,I_i)$ on
$\mathcal{Q}_{\mathrm{train}}$. All compared training methods use the same
subset for which $p_i$ differs from $q_i$ after Unicode normalization,
whitespace collapse, and case folding. D-OPCD trains from $(q_i,p_i)$; $I_i$
determines record eligibility but is not a loss target. Diffusion-DPO additionally
pairs the submitted image $I_i$ with a rejected image generated directly from
$q_i$ using the same seed. Vanilla SFT and D-OPSD train from the corresponding
$(q_i,I_i)$ records. The per-benchmark counts and filtering yields are reported in
Appendix~\ref{app:context_record_counts}.

\subsection{D-OPCD Implementation}
\label{app:dopcd_implementation}

The student receives $q_i$. For the teacher, the text encoder receives $q_i$
followed by a blank line, the label \texttt{Privileged generation context:},
and $p_i$. The student and teacher use the same frozen text encoder, with
maximum sequence lengths of 512 and 1024 tokens, respectively. Starting from
Gaussian noise, the student follows a four-point Euler rollout at
$t\in\{0,0.1,0.25,0.5\}$, with its final update ending at $t=1$.
The state is detached before each velocity comparison, so gradients update
the student at visited states without backpropagating through earlier rollout
steps. An EMA copy of the student, with decay $0.9999$, supplies the
teacher prediction under the additional context. We minimize the mean squared
velocity difference across the four states, as defined in
Section~\ref{sec:dopcd}; the teacher prediction is stop-gradient.

\subsection{Baseline Objectives}
\label{app:baseline_training}

\paragraph{Vanilla SFT.}
The student receives $q_i$, while the agent-submitted image $I_i$ supplies
the clean image latent for the standard flow-matching target. Each batch
samples a noise level from a logit-normal distribution with logit mean zero
and standard deviation one. No privileged prompt enters the model.

\paragraph{Diffusion-DPO.}
Both the chosen image $I_i$ and the same-seed rejected image are trained under
$q_i$. A frozen copy of the initial generator supplies reference losses, and
the chosen--rejected difference in model and reference flow-matching losses
forms the preference objective with $\beta=2500$. Two logit-normal noise
levels are sampled per pair.

\paragraph{D-OPSD.}
The student receives $q_i$ and visits states along its own sampling trajectory.
The EMA teacher, updated with decay $0.9999$, receives an embedding of
$(q_i,I_i)$ from Qwen3-VL-4B-Instruct and supplies a detached clean-output
endpoint target.
This baseline shares SFT's image-paired records, but uses on-policy student
states rather than the image-derived states used by SFT.

\subsection{Optimization and Checkpoint Selection}
\label{app:training_checkpoint_selection}

All methods start from Z-Image-Turbo and train LoRA adapters on its transformer
feed-forward projections and attention query, key, value, and output
projections. The common settings are rank 64, scale parameter $\alpha=128$,
zero LoRA dropout, 1024-pixel square images, and seed 42. We use AdamW with
$\beta_1=0.9$, $\beta_2=0.999$, zero weight decay, a constant learning rate
of $10^{-4}$ without warmup, gradient clipping at norm 1, and bfloat16
precision. Training runs for 2000 optimizer steps and saves a checkpoint every
200 steps. Direct evaluation uses eight sampling steps and guidance scale zero.

\paragraph{Compute resources.}
Model training and image-generation inference used NVIDIA A100 GPUs (80 GB
each). Training jobs allocated four GPUs; inference used one GPU per generation
worker, with multiple workers run in parallel where applicable.

Each saved checkpoint is evaluated on $\mathcal{Q}_{\mathrm{hold}}$; the
benchmark-specific selection metrics are defined in
Appendix~\ref{app:eval_protocol_details}. Table~\ref{tab:selected_training_checkpoints}
lists each benchmark and compared method using skill-equipped harness data,
with the reported optimizer step and held-out score. Hold
scores use the same benchmark-native metric used for checkpoint evaluation
and for the main evaluation tables.

\begin{table*}[t]
\centering
\small
\renewcommand{\arraystretch}{1.2}
\setlength{\tabcolsep}{0pt}
\caption{Generator checkpoints and held-out scores by method and benchmark. All methods use records from $\mathcal{H}_{r^\star}$; scores use GenEval Overall, GenEval2 Soft-TIFA AM, WIScore, and R2I-Score (0--100).}
\label{tab:selected_training_checkpoints}
\begin{tabular*}{0.70\textwidth}{>{\raggedright\arraybackslash}p{0.34\textwidth}w{c}{0.17\textwidth}w{c}{0.19\textwidth}}
\toprule
\textbf{Benchmark} & \textbf{Step} & \textbf{Hold score} \\
\midrule
\rowcolor{blue!7}
\textbf{D-OPCD} & & \\
GenEval & 1800 & 86.31 \\
GenEval2 & 1600 & 82.38 \\
WISE & 1000 & 48.67 \\
R2I-Bench & 400 & 49.33 \\
\addlinespace[0.3em]
\rowcolor{blue!7}
\textbf{Vanilla SFT} & & \\
GenEval & 1000 & 82.98 \\
GenEval2 & 1600 & 82.13 \\
WISE & 200 & 50.67 \\
R2I-Bench & 600 & 51.30 \\
\addlinespace[0.3em]
\rowcolor{blue!7}
\textbf{Diffusion-DPO} & & \\
GenEval & 200 & 79.29 \\
GenEval2 & 1400 & 79.49 \\
WISE & 1800 & 49.33 \\
R2I-Bench & 1200 & 47.87 \\
\addlinespace[0.3em]
\rowcolor{blue!7}
\textbf{D-OPSD} & & \\
GenEval & 1200 & 79.64 \\
GenEval2 & 1800 & 75.29 \\
WISE & 400 & 51.33 \\
R2I-Bench & 200 & 49.05 \\
\bottomrule
\end{tabular*}
\end{table*}

\section{Experimental and Evaluation Details}
\label{app:eval_protocol_details}

\subsection{Task Set Allocations}
\label{app:data_splits}

Table~\ref{tab:task_split_counts} gives the disjoint task sets used for
harness evolution, selection, and final evaluation. GenEval, GenEval2, and
WISE use approximately 50/15/35 allocations; R2I-Bench uses a 780-task
experimental subset with its supplied split.

\begin{table}[t]
\centering
\small
\renewcommand{\arraystretch}{1.15}
\setlength{\tabcolsep}{2pt}
\caption{In-distribution task allocations. Ratios give the train/hold/evaluation shares of the task pool actually used.}
\label{tab:task_split_counts}
\begin{tabular*}{\columnwidth}{@{\extracolsep{\fill}}lrrrrl@{}}
\noalign{\hrule height 0.9pt}
\textbf{Benchmark} & \textbf{Pool} & \textbf{$|\mathcal{Q}_{\mathrm{train}}|$} & \textbf{$|\mathcal{Q}_{\mathrm{hold}}|$} & \textbf{$|\mathcal{Q}_{\mathrm{eval}}|$} & \textbf{Ratio (\%)} \\
\hline
GenEval & 553 & 276 & 83 & 194 & 49.9/15.0/35.1 \\
GenEval2 & 800 & 400 & 120 & 280 & 50/15/35 \\
WISE Verified & 1000 & 500 & 150 & 350 & 50/15/35 \\
R2I-Bench subset & 780 & 480 & 100 & 200 & 61.5/12.8/25.6 \\
\noalign{\hrule height 0.9pt}
\end{tabular*}
\end{table}

The first three splits use seed 42; GenEval2 and WISE preserve earlier
task assignments, while R2I-Bench uses its supplied split.

\subsection{Replay Data Construction and Counts}
\label{app:context_record_counts}

Frozen replay on $\mathcal{Q}_{\mathrm{train}}$ yields one submitted
$(q_i,p_i,I_i)$ record per task. Table~\ref{tab:replay_record_counts}
reports the records retained before and after excluding normalized
$p_i=q_i$ pairs. D-OPCD, Diffusion-DPO, Vanilla SFT, and D-OPSD use the
prompt-changed tasks, with method-specific supervision. The replay uses the
selected skill-equipped harness version in
Table~\ref{tab:selected_harness_snapshots}.

\begin{table}[t]
\centering
\small
\renewcommand{\arraystretch}{1.15}
\setlength{\tabcolsep}{3pt}
\caption{Replay-record yields from the selected skill-equipped harness $\mathcal{H}_{r^\star}$. The last column gives the prompt-changed tasks used by all four compared training methods.}
\label{tab:replay_record_counts}
\begin{tabular*}{\columnwidth}{@{\extracolsep{\fill}}llrrr@{}}
\noalign{\hrule height 0.9pt}
\textbf{Benchmark} & \textbf{Source} & \textbf{Records} & \textbf{$p_i=q_i$} & \textbf{Changed} \\
\hline
GenEval & $\mathcal{H}_{r^\star}$ & 276 & 0 & 276 \\
GenEval2 & $\mathcal{H}_{r^\star}$ & 400 & 0 & 400 \\
WISE & $\mathcal{H}_{r^\star}$ & 500 & 10 & 490 \\
R2I-Bench & $\mathcal{H}_{r^\star}$ & 480 & 88 & 392 \\
\noalign{\hrule height 0.9pt}
\end{tabular*}
\end{table}

\subsection{Context Effectiveness and Diversity Diagnostics}
\label{app:experience_diagnostics}

\paragraph{Improvement-filtered pairs.}
The default set retains records from the selected skill-equipped harness
when the submitted prompt differs from the original query after
normalization. The improvement filter instead starts from all saved
records, including unchanged prompts. For each task, we pair the saved
image from the harness's selected attempt with the first image generated
directly from $q_i$. We take the requirement checklist $\mathcal{C}_i$
from the selected harness attempt and score both saved images against
that same checklist using the harness's internal verifier. Previously
recorded answers are reused when the checklist matches; otherwise, the
direct-query image is verified against $\mathcal{C}_i$. If $V_i(I)$ is
the number of passed checks for image $I$, we retain $(q_i,p_i)$ iff
$V_i(I_i^{p})>V_i(I_i^{q})$. Missing or invalid verifier responses are
excluded. This comparison requires neither new image generation nor
external benchmark scores. Since the saved images may use different
seeds, the filter selects observed verifier improvements rather than
estimating a seed-controlled prompt effect. It selects 43, 264, 319,
and 314 pairs on GenEval, GenEval2, WISE, and R2I-Bench, respectively.
Three R2I-Bench pairs have $p_i=q_i$ but satisfy the image-level
criterion.

\paragraph{Diversity-adapted sampling.}
When the Skill router selects applicable Skills, the default harness
combines their guidance with $q_i$ in one initial prompt rewrite. To
broaden the contexts produced by the same Skills, the diversity adapter
acts at this step, before the first image-generation attempt. It
samples one of four expression strategies at random for each task with
selected Skills. \emph{Local edit} preserves the query's wording and
order where possible while adding relevant guidance. \emph{Restate}
rewrites the request in new
words. \emph{Integrate} weaves guidance into a single coherent image
description. \emph{Separate} states the original request first and
places guidance in separate sentences. All four instruct the rewriter
to preserve the task requirements rather than create variety by
changing the requested content. Table~\ref{tab:adapter_strategies}
gives the exact strategy instructions, and
Figure~\ref{fig:prompt-diversity-adapter} shows the shared rewrite
prompt that receives $q_i$, the selected Skill
instructions, and the sampled strategy.

If no Skill is selected, the first attempt uses $q_i$ directly. The
Skill library, router, internal verifier, and later refinement steps
remain the same, with at most five generation attempts per task. We
record the prompt of the agent's final selected image as $p_i$, which
need not be its first or last attempted prompt. After excluding
normalized $p_i=q_i$ cases, the adapted replay yields 276, 400, 491,
and 381 pairs on GenEval, GenEval2, WISE, and R2I-Bench. All three
context sets use the same D-OPCD objective and student/$[q_i;p_i]$
teacher interface.

\begin{table}[!htbp]
\centering
\small
\renewcommand{\arraystretch}{1.15}
\caption{Exact expression-strategy instructions inserted into the diversity-adapter rewrite template.}
\label{tab:adapter_strategies}
\begin{tabular}{@{}p{0.17\linewidth}p{0.78\linewidth}@{}}
\toprule
\textbf{Mode} & \textbf{Instruction} \\
\midrule
\texttt{local\_edit} & Keep the original request's wording and order where possible. Incorporate applicable Skill guidance through local edits or additions. \\
\texttt{restate} & Write the resulting prompt afresh in your own words rather than retaining the original sentence structure. Express the same request together with applicable Skill guidance. \\
\texttt{integrate} & Weave applicable Skill guidance directly into the description of the request, as one coherent image-generation prompt. \\
\texttt{separate} & State the original request first, then express applicable Skill guidance in separate sentences within the same final image-generation prompt. Do not output explanations of your work. \\
\bottomrule
\end{tabular}
\end{table}

\begin{figure}[!htbp]
\centering
\begin{asepromptbox}{Diversity adapter --- initial prompt rewrite}
Produce the prompt for the first image-generation attempt. Use the supplied Skills as guidance for fulfilling the original request. Apply guidance only where it is relevant to that request, and preserve the request's meaning and requirements. If a Skill conflicts with the original request, the request takes priority. Return only the resulting prompt. Returning the original request unchanged is allowed.

### Initial-Prompt Skill Instructions
{SELECTED_SKILL_INSTRUCTIONS}

### Original Prompt
{ORIGINAL_QUERY}

### Expression strategy
{STRATEGY_INSTRUCTION}
This strategy changes expression, not the requested content. It does not authorize dropping applicable guidance or adding new requirements merely to create variety.

Return ONLY the final enhanced prompt.
\end{asepromptbox}
\caption{\textbf{Diversity-adapter rewrite template.} The strategy instruction is chosen from Table~\ref{tab:adapter_strategies}; all other text is shared across modes.}
\label{fig:prompt-diversity-adapter}
\end{figure}

\paragraph{Paired task effectiveness.}
For each retained $(q_i,p_i)$, we generate one image from $q_i$ and one
from $p_i$ using the same initial generator, seed, sampling settings,
and task. Both images are scored against the original $q_i$ and its
benchmark annotations. Let $S_b(q)$ and $S_b(p)$ denote the
benchmark-native aggregate scores for setting $b$ on exactly its
retained tasks, expressed on a 0--100 scale. Table~\ref{tab:experience_effectiveness}
reports $\Delta_b=S_b(p)-S_b(q)$ in percentage points,
$100\Delta_b/S_b(q)$ as a relative percentage, and the percentage of
individual tasks whose $p_i$ image scores strictly above its paired
$q_i$ image. GenEval uses Overall and GenEval2 uses Soft-TIFA AM.
The same-seed external evaluation is distinct from the agent's
internal-verifier criterion used for filtering.

\begin{table*}[t]
\centering
\scriptsize
\renewcommand{\arraystretch}{1.16}
\setlength{\tabcolsep}{3pt}
\caption{Effectiveness of the three training-pair settings on their retained
train tasks. $S(q)$ and $S(p)$ use the original query as the scoring target:
GenEval Overall, GenEval2 Soft-TIFA AM, WIScore, or R2I-Score, respectively.
$\Delta=S(p)-S(q)$ is in percentage points; Rel.\ is $100\Delta/S(q)$, and
Win is the percentage of individual pairs with a strictly higher $p$-image
score. The external benchmark scorer evaluates both images in each pair.}
\label{tab:experience_effectiveness}
\begin{tabular*}{\textwidth}{@{\extracolsep{\fill}}llrrrrrr@{}}
\noalign{\hrule height 0.9pt}
\textbf{Benchmark} & \textbf{Setting} & \textbf{$N$} & \textbf{$S(q)$} &
\textbf{$S(p)$} & \textbf{$\Delta$} & \textbf{Rel.\ (\%)} & \textbf{Win (\%)} \\
\hline
GenEval & Default & 276 & 73.76 & 85.71 & +11.95 & +16.20 & 19.57 \\
 & Improvement filter & 43 & 12.50 & 78.99 & +66.49 & +531.94 & 69.77 \\
 & Diversity adapter & 276 & 73.76 & 81.84 & +8.08 & +10.95 & 16.30 \\
\hline
GenEval2 & Default & 400 & 77.43 & 85.74 & +8.31 & +10.73 & 70.75 \\
 & Improvement filter & 264 & 73.37 & 85.35 & +11.98 & +16.33 & 78.79 \\
 & Diversity adapter & 400 & 77.43 & 86.76 & +9.33 & +12.06 & 72.75 \\
\hline
WISE & Default & 490 & 43.74 & 80.00 & +36.26 & +82.89 & 38.78 \\
 & Improvement filter & 319 & 29.92 & 79.91 & +49.99 & +167.12 & 51.72 \\
 & Diversity adapter & 491 & 43.87 & 80.67 & +36.80 & +83.90 & 40.33 \\
\hline
R2I-Bench & Default & 392 & 42.58 & 77.17 & +34.59 & +81.23 & 75.00 \\
 & Improvement filter & 314 & 37.87 & 76.43 & +38.56 & +101.83 & 79.62 \\
 & Diversity adapter & 381 & 42.48 & 75.79 & +33.31 & +78.43 & 74.54 \\
\noalign{\hrule height 0.9pt}
\end{tabular*}
\end{table*}

\paragraph{Context-text diversity.}
We jointly fit character 3--5-gram TF--IDF features to all three
settings within each benchmark. We measure normalized character Vendi
score~\citep{friedman2023vendi} (Vendi divided by sample size), mean
within-set pairwise cosine similarity, and corpus
Distinct-1~\citep{li-etal-2016-diversity} (unique words divided by
all words). Mean prompt length aids interpretation. To reduce
the effect of cohort size, Table~\ref{tab:experience_diversity} uses
equal-size subsamples, fixed by the smallest setting in each benchmark:
43/264/319/314 for GenEval/GenEval2/WISE/R2I-Bench. For larger
settings, it averages 100 subsamples drawn with seed 42. These are
text-diversity proxies rather than image-diversity scores.
The R2I-Bench improvement-filtered pool includes three unchanged
$p_i=q_i$ pairs; we retain them to match the model-training data.

A separate earlier matched-400 GenEval2 audit compared the default
SkillOnly contexts with those from the diversity adapter, fitting
TF--IDF jointly to those two corpora. It found normalized lexical Vendi
scores of 0.7402 and 0.7921, respectively, and Distinct-2 values of
0.3533 and 0.3862. Because this two-setting word-level audit used a
different feature space and sample protocol, its numbers are not inserted
into Table~\ref{tab:experience_diversity}.

\begin{table*}[t]
\centering
\scriptsize
\renewcommand{\arraystretch}{1.16}
\setlength{\tabcolsep}{3pt}
\caption{Text diversity of the retained privileged contexts $p_i$.
Normalized character Vendi and Distinct-1 are higher for more varied
contexts; mean pairwise character TF--IDF cosine is lower. Pool $N$ is
the number of retained pairs, and Eval $n$ is the equal-size subsample.
Length is the mean number of words.}
\label{tab:experience_diversity}
\begin{tabular*}{\textwidth}{@{\extracolsep{\fill}}llrrrrrr@{}}
\noalign{\hrule height 0.9pt}
\textbf{Benchmark} & \textbf{Setting} & \textbf{Pool $N$} & \textbf{Eval $n$} &
\textbf{Char Vendi $\uparrow$} & \textbf{Char cosine $\downarrow$} &
\textbf{Distinct-1 $\uparrow$} & \textbf{Length} \\
\hline
GenEval & Default & 276 & 43 & 0.7754 & 0.1519 & 0.1768 & 96.4 \\
 & Improvement filter & 43 & 43 & 0.7864 & 0.1431 & 0.1767 & 116.7 \\
 & Diversity adapter & 276 & 43 & 0.7841 & 0.1459 & 0.1817 & 83.5 \\
\hline
GenEval2 & Default & 400 & 264 & 0.4914 & 0.1374 & 0.0483 & 189.3 \\
 & Improvement filter & 264 & 264 & 0.4816 & 0.1386 & 0.0476 & 197.4 \\
 & Diversity adapter & 400 & 264 & 0.5664 & 0.1094 & 0.0590 & 111.0 \\
\hline
WISE & Default & 490 & 319 & 0.7583 & 0.0796 & 0.1680 & 112.3 \\
 & Improvement filter & 319 & 319 & 0.7339 & 0.0856 & 0.1592 & 122.9 \\
 & Diversity adapter & 491 & 319 & 0.8133 & 0.0610 & 0.1940 & 84.7 \\
\hline
R2I-Bench & Default & 392 & 314 & 0.7668 & 0.0708 & 0.1183 & 143.7 \\
 & Improvement filter & 314 & 314 & 0.7582 & 0.0721 & 0.1135 & 153.3 \\
 & Diversity adapter & 381 & 314 & 0.7946 & 0.0632 & 0.1265 & 123.5 \\
\noalign{\hrule height 0.9pt}
\end{tabular*}
\end{table*}

Character n-grams probe repetition in the prompt wording that the
adapter is designed to reduce. A word unigram--bigram audit agrees on
GenEval2, WISE, and R2I-Bench, but its GenEval differences are small
and mixed: normalized word Vendi is 0.9221 for the default and 0.9205
for the adapter, while Distinct-2 is 0.5469 and 0.5430, respectively.

\subsection{Evaluation Metrics and Protocols}

For in-distribution evaluation, each method generates one image per task
on $\mathcal{Q}_{\mathrm{eval}}$. We report each benchmark by its native
metric under its official evaluation protocol. GenEval~\citep{Ghosh_2023}
uses the Overall score from its detector-based checker; GenEval2~\citep{kamath2025geneval2}
uses prompt-level Soft-TIFA AM; WISE~\citep{niu2025wise} uses WiScore under
the WISE\_Verified binary protocol; and R2I-Bench~\citep{chen2025r2ibench}
uses R2I-Score from its checklist-based judge. Checkpoint candidates are
evaluated on $\mathcal{Q}_{\mathrm{hold}}$, with Overall
score for GenEval, Soft-TIFA AM for GenEval2, WiScore for WISE, and
R2I-Score for R2I-Bench. Reported hold scores use these same
evaluation metrics.

T2I-CompBench++~\citep{huang2025compbenchpp} is evaluated separately on
2,100 prompts from the seven categories in Table~\ref{tab:ood_generalization}
of its official \emph{validation} split, using three images per prompt with
seeds 42--44 and reporting scores by category. This OOD
set is not $\mathcal{Q}_{\mathrm{hold}}$ and is used neither for skill
evolution nor for model or harness selection. The base generator is
evaluated once, and each other row in Table~\ref{tab:ood_generalization}
reports one source-specific D-OPCD checkpoint, without averaging across
training sources.


\section{Use of AI Assistants}
\label{app:ai_use}

AI assistants were used in three roles. First, the experimental harness uses
GPT-5.6 Terra to plan and refine generation prompts, verify images, and
consolidate task experience into Insights and Skills
(Appendix~\ref{app:harness_implementation}). Second, outputs are scored with
the benchmark-native automated evaluators described in
Appendix~\ref{app:eval_protocol_details}, including the checklist-based judge
for R2I-Bench. Third, general-purpose AI writing tools were used to improve
wording and readability. They were not used to originate the research
questions, choose the experimental design, produce the reported results, or
draw the paper's conclusions. The authors reviewed and approved the
experimental design, analyses, results, and manuscript text.

\end{document}

%% file: math_commands.tex
\usepackage{amsmath,amsfonts,bm}

\def\eqref#1{equation~\ref{#1}}
\def\Eqref#1{Equation~\ref{#1}}

\def\1{\bm{1}}

\DeclareMathAlphabet{\mathsfit}{\encodingdefault}{\sfdefault}{m}{sl}
\SetMathAlphabet{\mathsfit}{bold}{\encodingdefault}{\sfdefault}{bx}{n}

